%% file: main.tex
\documentclass[accepted]{uai2026} 
                        
\usepackage[american]{babel}

\usepackage{natbib} 
\usepackage{mathtools} 
\usepackage{booktabs} 
\usepackage{tikz} 

\usepackage{physics}
\usepackage{subcaption}
\usepackage{multirow}
\usepackage{makecell}
\usepackage{algorithm}
\usepackage[noend]{algpseudocode}
\input{commands}

\newtheorem{proposition}{Proposition}

\title{
Simplex-to-Euclidean Bijection for Conjugate and \\ Calibrated Multiclass Gaussian Process Classification
}

\author[1]{\href{mailto:<bernardo.williamsmoreno@helsinki.fi>}{Bernardo Williams}{}}
\author[1]{Harsha Vardhan Tetali}
\author[1]{Arto Klami}
\author[1]{Marcelo Hartmann}
\affil[1]{%
    Department of Computer Science, University of Helsinki, Finland    
}
  
\begin{document}
\maketitle

\begin{abstract}
We propose a conjugate and calibrated Gaussian process (GP) model for multi-class classification by exploiting the geometry of the probability simplex. Our approach uses Aitchison geometry to map simplex-valued class probabilities to an unconstrained Euclidean representation, turning classification into a GP regression problem with fewer latent dimensions than standard multi-class GP classifiers. This yields conjugate inference and reliable predictive probabilities without relying on distributional approximations in the model construction. The method is compatible with standard sparse GP regression techniques, enabling scalable inference on larger datasets. Empirical results show well-calibrated and competitive performance across synthetic and real-world datasets.
\end{abstract}

\section{Introduction}\label{sec:intro}

Gaussian processes (GPs) \citep{williams2006gaussian} are attractive probabilistic models for supervised learning: they provide well-principled uncertainty quantification and often perform strongly in small-data regime. In multi-class classification with $K$ categories, however, the standard GP approach introduces a vector of latent functions (typically one per category) and maps them to class probabilities with a link function such as the softmax \citep{williams1998bayesian}. This yields a non-conjugate likelihood, so posterior inference and parameter learning need approximative techniques such as variational inference, Laplace approximation, or Markov chain Monte Carlo \citep{nickisch2008approximations,hensman2015mcmc, hernandez2016scalable}.

\begin{figure}[t]
\centering
\setlength{\tabcolsep}{3pt}
\renewcommand{\arraystretch}{1.05}
\newlength{\imgw}
\setlength{\imgw}{0.12\textwidth}
\newlength{\imgwide}
\setlength{\imgwide}{\dimexpr 3\imgw + 4\tabcolsep\relax}

\begin{tabular}{cccc}
\rotatebox{90}{ \textbf{Data}} &
\multicolumn{3}{c}{\includegraphics[width=\imgwide]{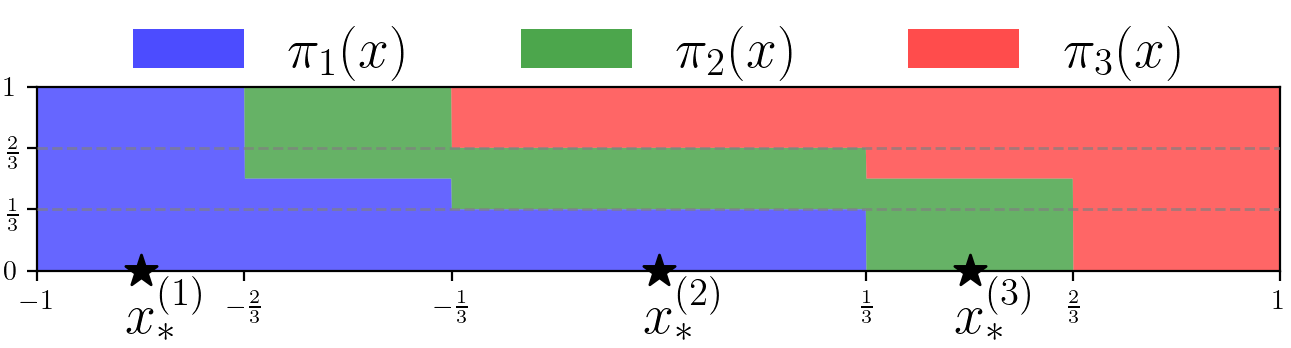}} \\
\hline
\rotatebox{90}{ \textbf{Likelihood}} &
\multicolumn{3}{c}{
\includegraphics[height=0.34\imgwide]{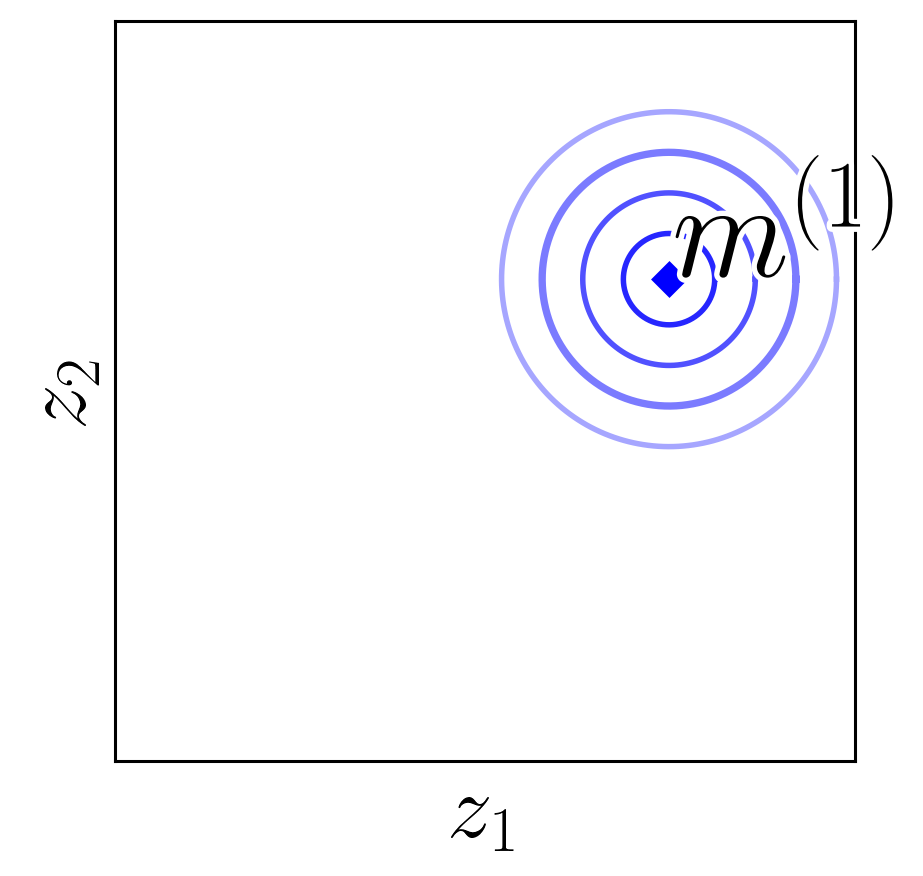} 
\includegraphics[height=0.34\imgwide]{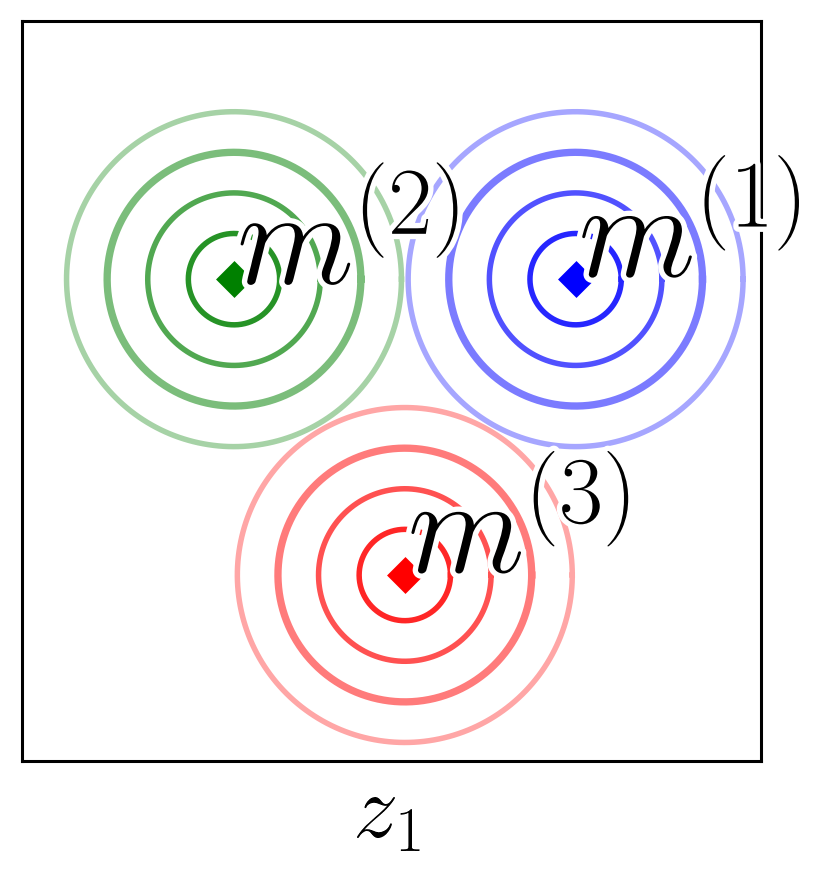} 
\includegraphics[height=0.34\imgwide]{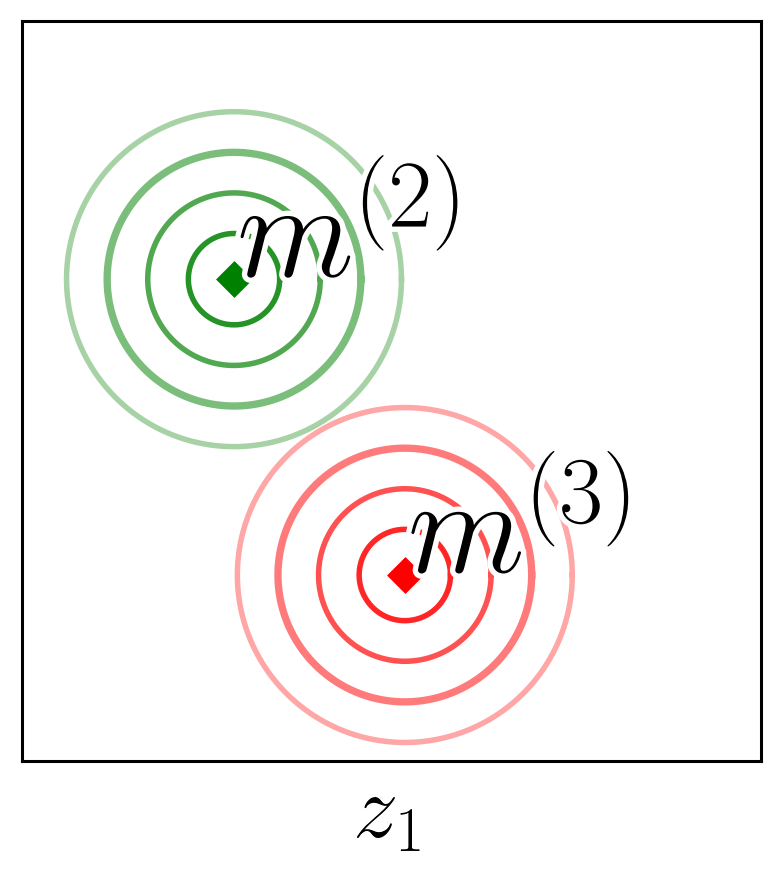} 
} \\
\hline
\multirow{3}{*}{\rotatebox{90}{\textbf{Posterior}}} &
\includegraphics[width=\imgw]{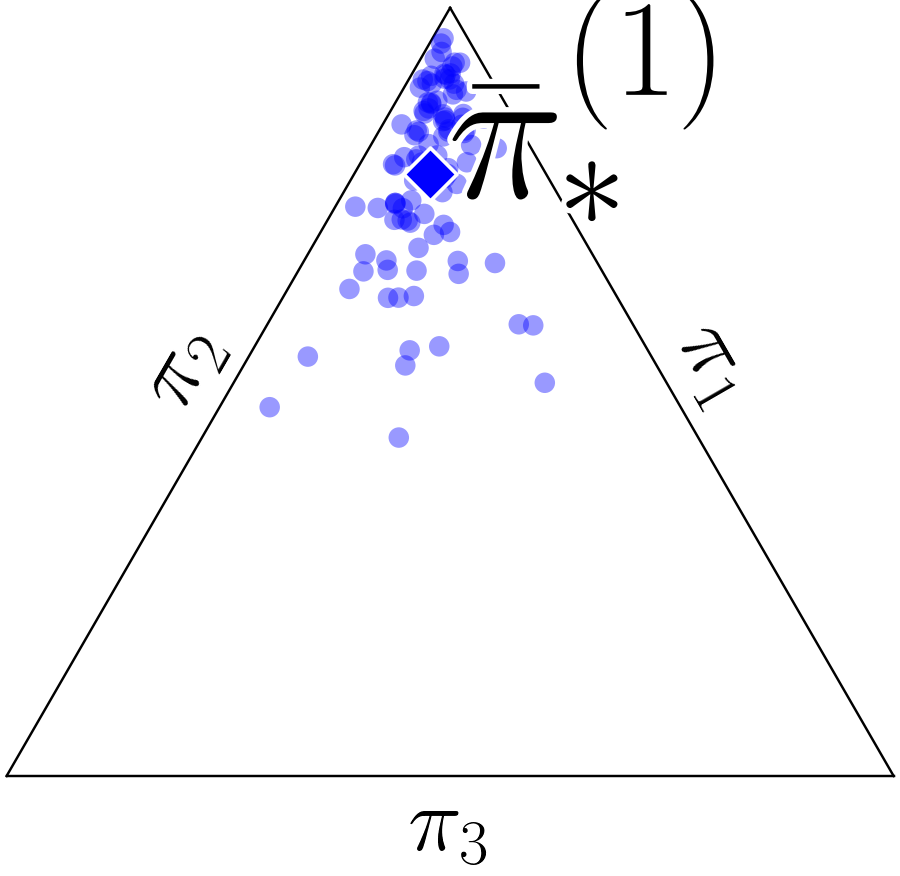} &
\includegraphics[width=\imgw]{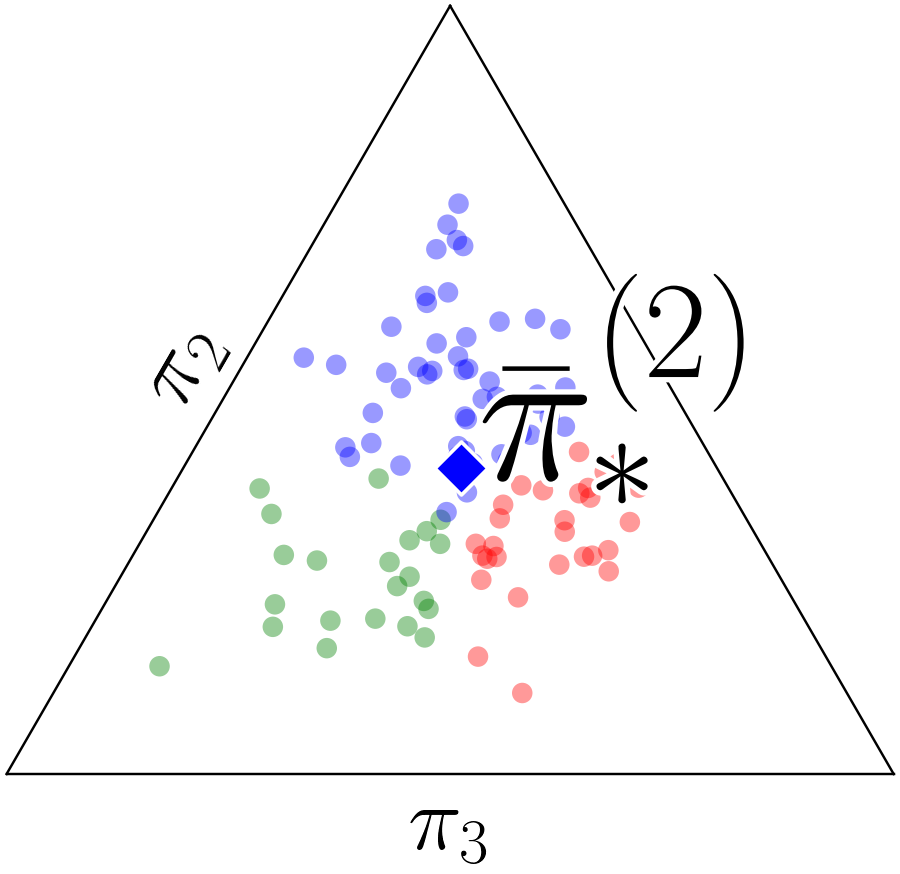} &
\includegraphics[width=\imgw]{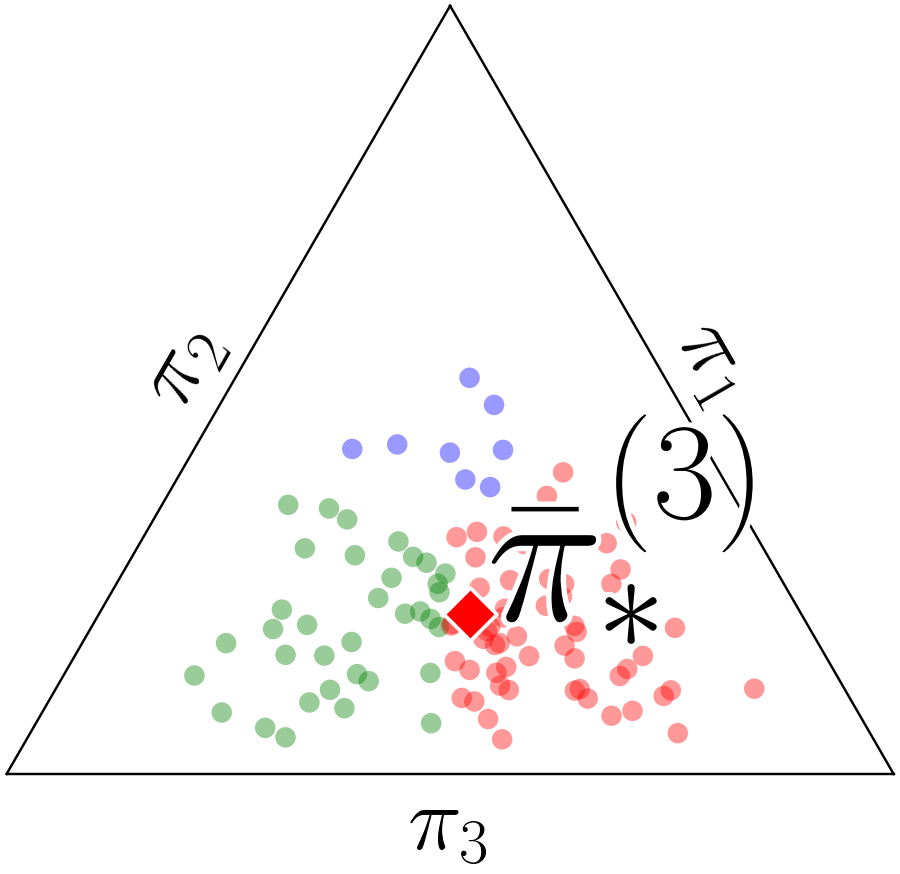} \\
& \multicolumn{3}{c}{\includegraphics[width=\imgwide]{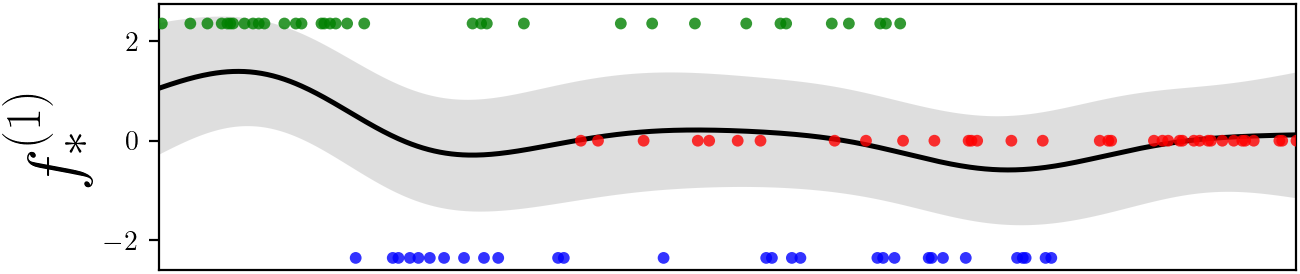}} \\
& \multicolumn{3}{c}{\includegraphics[width=\imgwide]{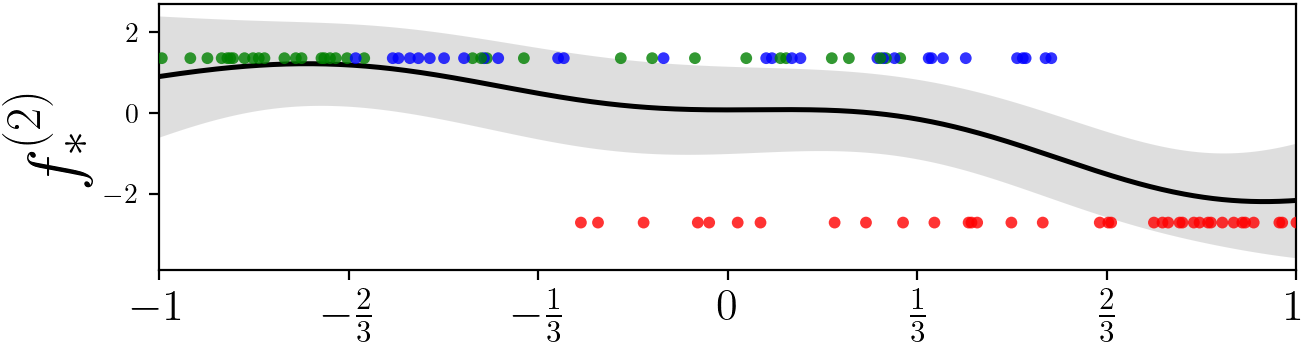}} \\
\end{tabular}
\caption{
Illustration of \ourmethod\ on a $K=3$ toy problem.
\textbf{Top (data):} Ground-truth class probabilities (blue, green and red), $\boldsymbol\pi(x)$ over $x\in[-1,1]$, and three test inputs $x_*^{(1:3)}$.
\textbf{Middle (likelihood):} Discrete labels are represented by Gaussian pseudo-observations centered at $\boldsymbol m^{(k)}=\varphi(\boldsymbol\mu^{(k)})$ in Euclidean space (Sec.~\ref{sec:method}), where $\boldsymbol\mu^{(k)}=\lambda\,\mathbf e_k+(1-\lambda)\tfrac{1}{K}\boldsymbol 1$.
The three panels visualize this class-weighted latent-space likelihood at $x_*^{(1:3)}$.
\textbf{Bottom (posterior):} For each $x_*^{(i)}$, we draw posterior samples of the latent GP, map them through $\varphi^{-1}$, and plot the resulting samples of $\boldsymbol\pi_*$. Each point is one sampled probability vector (color = $\arg\max$ class) and the average is $\bar{\boldsymbol\pi}_*^{(i)}$.
The two lowermost panels show the posterior over the two latent GP coordinates across $x\in[-1,1]$, with training points overlaid.
}
\label{fig:visual-abstract}
\end{figure}

For many applications, two desiderata are especially important: (i) sharp and reliable calibration of predictive probabilities \citep{niculescu2005predicting, guo2017calibration}, and (ii) inference procedures that are exact given the model (i.e., conjugate), rather than relying on variational approximations in the model construction. From the perspective of inference, 
notable effort has been made to retain conjugacy in various forms. The simplest approaches model discrete labels via direct GP regression \citep{frohlich2013large}, whereas \citet{wenzel2019efficient} and \citet{galyfajou2022latent} introduced auxiliary variables to obtain conditional conjugacy and enable Gibbs sampling from conditionally conjugate marginals.
Dirichlet-based GP classification (GPD) by \citet{milios2018dirichlet}
gets closest to offering a fully conjugate solution, avoiding auxiliary variables by using a gamma reparameterization for a Dirichlet distribution. However, they still need to approximate the gamma distributions with log-normal distributions for conjugacy.

In practice, many of these methods exhibit poor calibration. Direct GP regression is reported to be very poorly calibrated without additional scaling \citep{milios2018dirichlet}, and while robust variants based on Heaviside likelihoods improve robustness to outliers \citep{hernandez2011robust, villacampa2021multi} they remain poorly calibrated. Overall, variational GP classification \citep{Hensman2015} and GPD typically provide the best calibration in practice.

In this work, we present a method that is fully conjugate without relying on any augmentation or approximations but retains good calibration, revisiting GP classification from the perspective of the geometry of probability vectors. Class probabilities live on the simplex, i.e., the set of nonnegative vectors in $\mathbb R^K$ whose entries sum to one. This space is naturally studied in compositional data analysis \citep{Aitchison1980, Aitchison1982, Egozcue2003} which has recently been used in generative modeling on the simplex \citep{Diederen2025, Williams2026, Chereddy2025}. Utilizing Aitchison geometry, we use the simplex-to-Euclidean bijection of \citet{Egozcue2003} to map probability vectors to an unconstrained Euclidean representation. This allows us to replace discrete labels with continuous Gaussian pseudo-observations in Euclidean space and reduce multi-class classification to GP regression in $D:=K-1$ latent dimensions.

The resulting model is fully conjugate in Euclidean space: posterior inference and learning can be performed exactly using the standard GP marginal likelihood, with the usual exact GP regression cost of $\mathcal O(N^3)$; see Fig~\ref{fig:visual-abstract} for an illustration. 
Concretely, we use the Isometric log-ratio bijection of \citet{Egozcue2003} to define a GP prior over a $D$ dimensional Euclidean latent representation, and we train the model via standard GP regression machinery on Gaussian pseudo-observations derived from class labels. Compared to Dirichlet-based GP classification \citep{milios2018dirichlet}, this yields a conjugate model without introducing a distributional approximation in the construction and reduces the number of latent processes from $K$ to $D$. Because the resulting learning problem is GP regression in a transformed space, it can leverage essentially any sparse GP regression technique (e.g., inducing-point methods) when scaling beyond the exact $\mathcal O(N^3)$ setting \citep{titsias2009variational, Hensman2013}. 

Our main empirical observation is that there are very large differences in how well GP classifiers are calibrated, with ours and GPD by \citet{milios2018dirichlet} being consistently the best methods with a clear gap between them and the rest. In other words, we provide a new model that is fully conjugate and empirically performs extremely well, but the empirical improvement over the strong baseline of GPD is small.

\section{Background}

\paragraph{Gaussian Process classification}
We consider multi-class classification with data $\mathcal D=\{(\boldsymbol{x}_n,c_n)\}_{n=1}^N$, where $\boldsymbol{x}_n\in\mathbb{R}^P$ and $c_n\in\{1,\dots,K\}$. Let $D:=K-1$ denote the intrinsic dimension of the simplex. We denote the (closed) probability simplex 
$\Delta^{D}:=\{\boldsymbol\pi\in\mathbb R_{\ge 0}^{K}:\;\sum_{k=1}^{K}\pi_k=1\}$ and its interior
$\mathring\Delta^{D}:=\{\boldsymbol\pi\in\mathbb R_{> 0}^{K}:\;\sum_{k=1}^{K}\pi_k=1\}$. The goal is to estimate the unknown class probabilities
\[
\boldsymbol\pi(\boldsymbol{x})=(\pi_1(\boldsymbol{x}),\dots,\pi_K(\boldsymbol{x}))\in\Delta^{D},
\
c\mid \boldsymbol{x} \sim \mathrm{Cat}\!\bigl(\boldsymbol\pi(\boldsymbol{x})\bigr).
\]
That is, $\boldsymbol\pi(\cdot)$ is an unknown function that maps input covariates $\boldsymbol{x}$ to a probability vector for the $K$ categories, and we aim to learn/estimate this mapping from data.
A standard GP classifier introduces latent functions and a link function, typically
\begin{equation}
\mathbf f(\boldsymbol{x})=(f_1(\boldsymbol{x}),\dots,f_K(\boldsymbol{x}))\in\mathbb R^K,
\
\boldsymbol\pi(\boldsymbol{x})=\mathrm{softmax}\!\bigl(\mathbf f(\boldsymbol{x})\bigr),
\label{eq:softmax}
\end{equation}
with independent GP priors on the components,
\[
f_k(\cdot)\sim \mathcal{GP}\!\bigl(0,\,K_{\theta}(\cdot,\cdot)\bigr),\qquad k\in\{1,\dots,K\},
\]
where $K_{\theta}$ is a kernel with hyperparameters $\theta$. The posterior is non-conjugate due to the softmax likelihood, so inference is usually approximate, e.g., scalable variational inference with inducing points \citep{Hensman2015}.

\paragraph{Dirichlet-based Gaussian process classification (GPD)}
\citet{milios2018dirichlet} proposed Dirichlet-based Gaussian process classification (GPD) as a conjugate and calibrated approach to multi-class classification. The model places a Dirichlet distribution on the probability vector,
\[
\boldsymbol\pi(\boldsymbol{x})\mid \boldsymbol{x} \sim \mathrm{Dir}\!\bigl(\boldsymbol\alpha(\boldsymbol{x})\bigr),
\qquad
c\mid \boldsymbol{x} \sim \mathrm{Cat}\!\bigl(\boldsymbol\pi(\boldsymbol{x})\bigr).
\]
Given an observed class $c=k$, GPD uses a smoothed target mean and a concentration vector
\[
\boldsymbol\mu^{(k)} := \varepsilon\,\mathbf e_k + (1-\varepsilon)\,\tfrac{1}{K}\boldsymbol 1,
\qquad
\boldsymbol\alpha := \alpha_0\,\boldsymbol\mu^{(k)},
\]
with $\alpha_0 := 1 + K\alpha_{\varepsilon}$ and $\varepsilon := (1+K\alpha_{\varepsilon})^{-1}$ for some $\alpha_\varepsilon>0$. This yields $\alpha_k=1+\alpha_{\varepsilon}$ and $\alpha_j=\alpha_{\varepsilon}$ for $j\neq k$. As $\alpha_{\varepsilon}\to 0$, the Dirichlet distribution concentrates on the corresponding one-hot vertex $\mathbf e_k$: its marginal variances shrink and mass moves to the simplex boundary. In practice, the authors suggest $\alpha_{\varepsilon}$ is chosen by validation.

This method decomposes the Dirichlet distribution into independent Gamma distributions $g_j\stackrel{\text{ind}}{\sim}\mathrm{Gamma}(\alpha_j,1)$, so that $\pi_j=g_j/\sum_{\ell} g_{\ell}$ is distributed $\mathrm{Dir}(\boldsymbol{\alpha})$. Matching the first two moments of $g_j$ with a log-normal approximation $\mathrm{Lognormal}(\tilde{y}_j,\tilde{\sigma}_j^2)$ yields
\begin{equation}
\begin{aligned}
\tilde{\sigma}_j^2&=\log\!\left(1+\tfrac{1}{\alpha_j}\right),
\qquad
\tilde{y}_j=\log \alpha_j-\tfrac{1}{2}\tilde{\sigma}_j^2.
\end{aligned}    
\label{eq:lognorm}
\end{equation}

Equivalently, in log-space,
\[
\log g_j \approx \mathcal N(\tilde{y}_j,\tilde{\sigma}_j^2),
\]
so the problem becomes conjugate after the log transform: one fits $K$ independent GPs, each $f_j$ to pseudo-observations $\tilde{y}_{nj}$ with a heteroscedastic Gaussian likelihood whose noise variances $\tilde{\sigma}_{nj}^2$ depend on the (label-dependent) Dirichlet parameters $\alpha_{nj}$ for $n=1,..,N$. This view is close in spirit to our approach by also leveraging Gaussian regression in a transformed space.

At a test input $x_*$, letting $\mathbf f_*\in\mathbb R^K$ denote the latent GP, GPD approximates predictive probabilities by Monte Carlo,
\[
\begin{aligned}
\mathbb E[\boldsymbol\pi_* \mid \boldsymbol{x}_*,\mathcal D]
&\approx \frac{1}{S}\sum_{s=1}^S \mathrm{softmax}\bigl(\mathbf f_*^{(s)}\bigr),\\
\mathbf f_*^{(s)} &\sim p(\mathbf f_* \mid \boldsymbol{x}_*,\mathcal D).
\end{aligned}
\]

\section{Method}\label{sec:method}

We propose a conjugate and calibrated GP classification model by mapping probability vectors on the simplex to Euclidean space, and converting classification into regression by assigning each class a latent target location and learning a GP that interpolates these targets. The construction follows the same high-level idea as GPD, but it avoids the log-normal approximation and uses $D:=K-1$ latent dimensions, one less than GPD.

GPD models $\boldsymbol\pi\mid \boldsymbol{x}\sim \mathrm{Dir}(\boldsymbol\alpha(\boldsymbol{x}))$ with a GP. We instead, model class probabilities through a latent Euclidean representation of the open simplex. We choose a bijection $\varphi:\mathring\Delta^D\to\mathbb R^D$ and learn a latent GP for $\mathbf f(\boldsymbol{x})\in\mathbb R^D$, and eventually the class probabilities are estimated by mapping back to the simplex with
\[
\tilde{ \boldsymbol\pi}(\boldsymbol{x}) := \varphi^{-1}\!\bigl(\mathbf f(\boldsymbol{x})\bigr)\in\mathring\Delta^{D}.
\]
To train the model, each discrete label $c=k$ is associated with a class target $\boldsymbol m^{(k)}\in\mathbb R^D$ obtained by mapping a smoothed one-hot vector $\boldsymbol\mu^{(k)}$ through $\varphi$. We then fit $\mathbf f$ by GP regression to these latent targets under a Gaussian likelihood with a fixed variance $\sigma^2$. This yields conjugate inference. The precise choices of $\varphi$, $\boldsymbol\mu^{(k)}$, $\boldsymbol m^{(k)}$, and $\sigma^2$ are given next.

\begin{figure}
    \centering
    \includegraphics[width=0.9\linewidth]{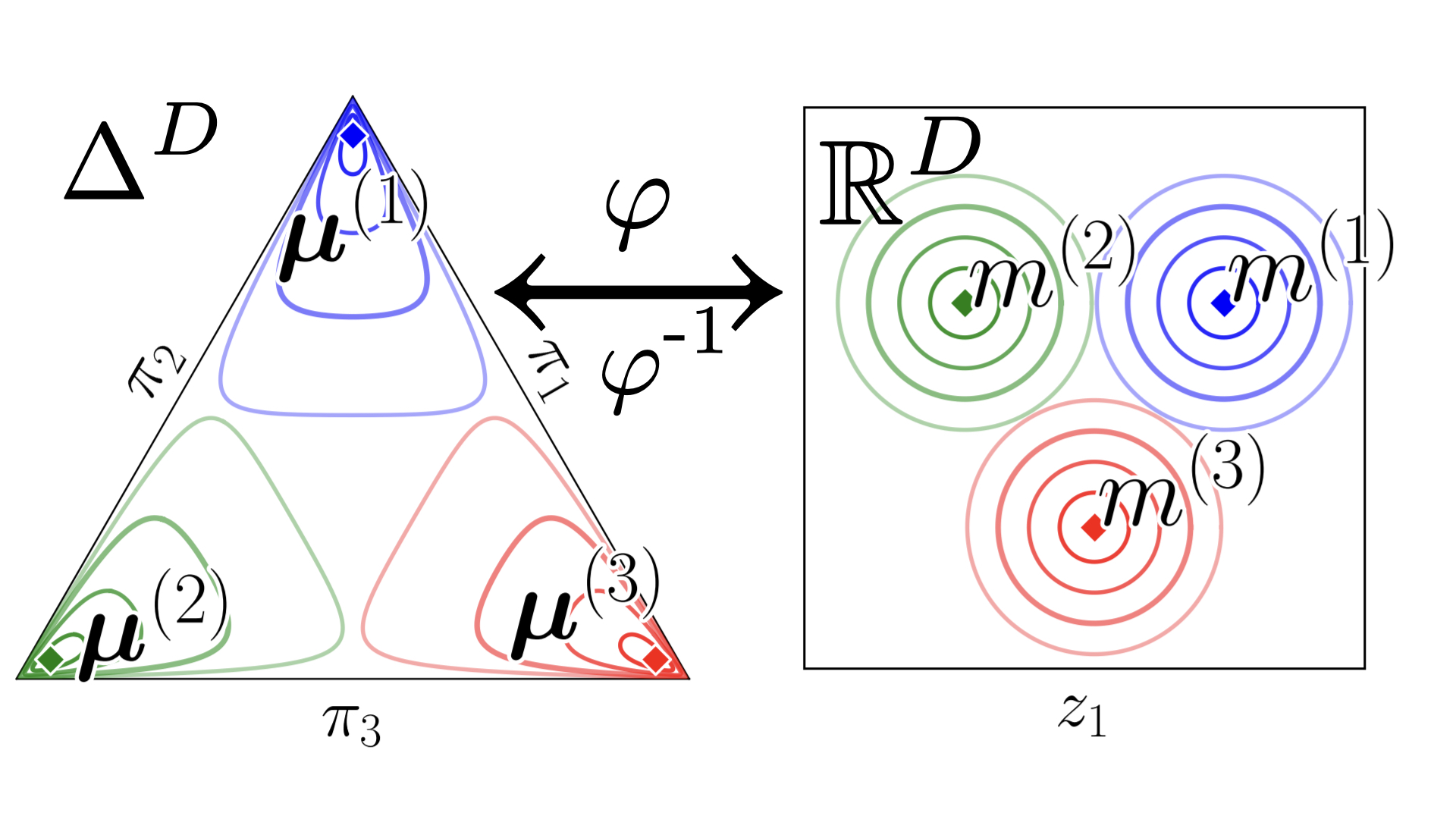}
    
    \caption{
    Likelihood in simplex and Euclidean spaces. The likelihood of observing classes $1$, $2$, or $3$ at input $x$ is a Gaussian mixture with  modes at $\boldsymbol m^{(k)}=\varphi(\boldsymbol\mu^{(k)})$ in Euclidean space (right), which induces a pushforward likelihood on the simplex (left). The figure shows the likelihood for $\boldsymbol{\pi}(x) = (1/3,1/3,1/3)$.}
    \label{fig:likelihood}
\end{figure}

\subsection{ILR-Gaussian Process} 

We now make the construction concrete. We (i) choose $\varphi$ to be the isometric logratio (ILR) bijection and (ii) specify how each class label is embedded as a latent target with an adequate amount Gaussian noise. We build on the idea recently proposed by \citet{Williams2026} where a similar interpolation and simplex-to-Euclidean bijections was used for generative modeling of discrete data using flows. 

Given a label $c=k$, we map the one-hot vector to the simplex interior with the interpolation
\[
\boldsymbol\mu^{(k)} := \lambda\,\mathbf e_k + (1-\lambda)\,\tfrac{1}{K}\boldsymbol 1 \in \mathring\Delta^{D},
\qquad \lambda\in(0,1),
\]
define the corresponding latent target $\boldsymbol m^{(k)}:=\varphi\!\bigl(\boldsymbol\mu^{(k)}\bigr)$, and use the latent-space likelihood
\begin{equation}
\boldsymbol z\mid c=k
\sim
\mathcal N\!\bigl(\boldsymbol m^{(k)},\,\sigma^2\mathbf I_D\bigr),    
\label{eq:lik}
\end{equation}
where $\sigma^2$ is fixed (and chosen by a principled rule; see Proposition~\ref{prop:sigma_bound}). Mapping back through $\varphi^{-1}$ induces the simplex-valued pushforward likelihood (see Fig.~\ref{fig:likelihood} for an illustration),
\[
\boldsymbol\pi\mid c=k\sim (\varphi^{-1})_{\#}\,\mathcal N\!\bigl(\boldsymbol m^{(k)},\,\sigma^2\mathbf I_D\bigr).
\]
Training reduces to GP regression with a GP prior in $\mathbb{R}^D$ and the likelihood given in Equation~\ref{eq:lik}. The  posterior is Gaussian, inference is conjugate on $\mathbb{R}^D$ and class probabilities are generated by mapping to $\mathring\Delta^D$ with $\varphi^{-1}$ (full details provided in Sec.~\ref{sec:train}).
The main design choices are the mapping $\varphi$ and the noise level $\sigma^2$, which controls the class overlap in the latent space. In our experiments, we select $\lambda$ by validation.

\paragraph{ILR map and Aitchison geometry}
Here, we briefly introduce Aitchison geometry and the associated ILR bijection, which maps the open simplex to Euclidean space. This framework provides a principled notion of norms and distances between probability vectors on the simplex, and the ILR map gives Euclidean coordinates that preserve these distances. We will later use this geometry to define a suitable noise level in the likelihood (Proposition~\ref{prop:sigma_bound}) by quantifying the separation between the latent class targets $\boldsymbol m^{(k)}$ in a consistent way. We equip the open simplex with the Aitchison inner product \citep{Aitchison1982}. For $\boldsymbol x,\boldsymbol y\in\mathring\Delta^{D}$,
\begin{equation*}
\langle \boldsymbol{x}, \boldsymbol{y} \rangle_A := \frac{1}{2K} \sum_{i,j=1}^{K}
\log\frac{x_i}{x_j}\log\frac{y_i}{y_j}.
\end{equation*}
This inner product induces the norm $\|\boldsymbol x\|_A:=\sqrt{\langle \boldsymbol x,\boldsymbol x\rangle_A}$ and the Aitchison distance
\[
d_A(\boldsymbol x,\boldsymbol y)
:= \sqrt{\frac{1}{2K}\sum_{i,j=1}^K
\left(\log\frac{x_i}{x_j}-\log\frac{y_i}{y_j}\right)^2 }.
\]
The isometric logratio (ILR) transform \citep{Egozcue2003} provides explicit Euclidean coordinates that respect this geometry. Let $\boldsymbol{H}\in\mathbb R^{D\times K}$ be the Helmert matrix \citep{Lancaster1965} and define
\begin{equation} \label{eq:ilr}
    \begin{aligned}
    \varphi: \mathring\Delta^{D} \to \mathbb{R}^{D},\quad  &\boldsymbol{y}\mapsto \boldsymbol{z}= \boldsymbol{H} \log \boldsymbol{y}, \\
    \varphi^{-1}:  \mathbb{R}^{D} \to \mathring\Delta^{D}, \quad &\boldsymbol{z}\mapsto \boldsymbol{y} = \mathrm{softmax}\!\bigl( \boldsymbol{H}^\top \boldsymbol{z} \bigr),
    \end{aligned}
\end{equation}
where $\log$ is applied elementwise.
The ILR mapping is an isometry between $(\mathring\Delta^{D},d_A)$ and $(\mathbb R^{D},\|\cdot\|_2)$, so Euclidean distances in latent space correspond exactly to Aitchison distances on the simplex. This makes it convenient to place a Gaussian likelihood and a GP prior on $\boldsymbol z\in\mathbb R^D$ while preserving the simplex-geometry notion of separation between probability vectors. The complexity of the ILR-transform is linear on the number of classes $\mathcal{O}(K)$, due to the special form of the Helmert matrix. Other diffeomorphic log-ratio maps such as the additive log-ratio or multiplicative log-ratio can also be used \citep{Aitchison1981, Aitchison1982}, but these mappings depend on the order of the input, giving a different map for different permutations and are not isomorphic. 

\paragraph{How to choose $\sigma^2$?}
The variance $\sigma^2$ controls how much overlap there is between pseudo-observations of different classes in latent space.  We find an upper bound for the variance that limits the overlap to a given tolerance probability $\epsilon$. We can select any value of $\sigma$ between zero and the upper bound, but in practice choose $\sigma$ as the upper bound value, and a smaller value can be chosen by reducing the overlap tolerance. 

Marginalizing the (unknown) class label, our pseudo-observation model corresponds to a Gaussian mixture,
\begin{equation*}
\begin{aligned}
p(\boldsymbol z\mid \boldsymbol{x})
&= \sum_{k=1}^K p(\boldsymbol z,c=k\mid \boldsymbol{x}) \\
&= \sum_{k=1}^K p(c=k\mid \boldsymbol{x})\, p(\boldsymbol z\mid c=k,\boldsymbol{x}) \\
&= \sum_{k=1}^K \pi_k(\boldsymbol{x})\,\mathcal N\!\bigl(\boldsymbol z\mid \varphi(\boldsymbol\mu^{(k)}),\,\sigma^2\mathbf I_D\bigr).
\end{aligned}
\end{equation*}
To control component overlap, we use the fact that the ILR map $\varphi$ is an isometry: Euclidean distances between component means in latent space equal Aitchison distances on the simplex.
We define the (common) pairwise separation
\[
\delta := d_A\!\bigl(\boldsymbol\mu^{(k)},\boldsymbol\mu^{(\ell)}\bigr) = \|\boldsymbol m^{(k)}-\boldsymbol m^{(\ell)}\|_2,\qquad k\neq \ell,
\]
and derive Proposition~\ref{prop:sigma_bound} as a sufficient condition for negligible intersection between mixture components.

\begin{proposition}[Choice of $\sigma$ for negligible component intersection]\label{prop:sigma_bound}
Let $K\ge 2$.  Define, for each class $k\in\{1,\dots,K\}$,
\[
\boldsymbol m^{(k)} := \varphi\!\bigl(\boldsymbol\mu^{(k)}\bigr)\in\mathbb R^{D}.
\]
Let
\[
\delta := \|\boldsymbol m^{(k)}-\boldsymbol m^{(\ell)}\|_2, \quad k\neq \ell,
\]
which equals the corresponding Aitchison distance between the class centers on the simplex by the ILR isometry.
Let $\mathcal V_k:=\{z\in\mathbb R^{D}:\|z-\boldsymbol m^{(k)}\|\le \|z-\boldsymbol m^{(\ell)}\|\ \forall \ell\}$.
For any $\varepsilon\in(0,1)$, if
\begin{equation}\label{eq:sigma_bound}
\sigma \;\le\; \frac{\delta}{2\; z_{1-\varepsilon/D}},
\qquad z_{q}:=\Phi^{-1}(q),
\end{equation}
then for every $\boldsymbol{x}$ and every $k$,
\[
\mathbb P\!\big(\boldsymbol Z\notin \mathcal V_k \,\big|\, C=k,\boldsymbol{x}\big)\;\le\;\varepsilon,
\]
where $C$ denotes the class label and $\boldsymbol Z\sim\mathcal N(\boldsymbol m^{(k)},\sigma^2\mathbf I_D)$.
\end{proposition}

\subsection{Training and prediction} \label{sec:train}

\paragraph{Exact Gaussian process model (\ourmethod)}
We place a GP prior on a latent function $\mathbf f(\cdot)\in\mathbb R^{D}$ and use a Gaussian likelihood,
\[
\mathbf f(\cdot) \sim \mathcal{GP}(\mathbf 0, K_{\theta}),
\qquad
\boldsymbol z_i\mid \mathbf f(\boldsymbol{x}_i) \sim \mathcal N\!\bigl(\mathbf f(\boldsymbol{x}_i),\,\sigma^2\mathbf I_D\bigr).
\]
This is standard multi-output GP regression (with independent outputs), hence the posterior inference and predictions are closed-form. This is the exact (non-sparse) variant used in our experiments. We follow \citet{milios2018dirichlet} and share the kernel's parameters across all dimensions. We infer the kernel hyperparameters $\theta$ by maximizing the (Gaussian) marginal likelihood $p(\{\boldsymbol z_i\}_{i=1}^N\mid\{\boldsymbol{x}_i\}_{i=1}^N,\theta)$.
Concretely, letting $\mathbf X:=(\boldsymbol{x}_1,\ldots,\boldsymbol{x}_N)$ and $\mathbf{K}_{N}:=K_{\theta}(\mathbf X,\mathbf X)\in\mathbb R^{N\times N}$, the model factorizes over ILR coordinates $d\in\{1,\ldots,D\}$. Writing $\mathbf z^{(d)}:=(z_{1d},\ldots,z_{Nd})^\top$ and defining the shorthand $\|\mathbf v\|^2_{\mathbf A}:=\mathbf v^\top\mathbf A^{-1}\mathbf v$, the marginal log-likelihood decomposes as
\begin{equation*}
\log p\bigl(\boldsymbol z\mid \theta\bigr)
= -\tfrac12\sum_{d=1}^D \|\boldsymbol z^{(d)}\|^2_{(\mathbf{K}_{N}+\sigma^2\mathbf I_N)^{-1}}
- \tfrac{D}{2}\log\lvert\mathbf{K}_{N}+\sigma^2\mathbf I_N\rvert,
\end{equation*}
where we omit the additive constant $-\tfrac{ND}{2}\log(2\pi)$ since it does not affect optimization.
For a test input $\boldsymbol{x}_*$, with $\mathbf k_*:=K_{\theta}(\mathbf X,\boldsymbol{x}_*)$ and $k_{**}:=K_{\theta}(\boldsymbol{x}_*,\boldsymbol{x}_*)$, the predictive latent is
\begin{equation*}
\begin{aligned}
f_*^{(d)}\mid \boldsymbol z^{(d)},\mathbf X &\sim \mathcal N\!\bigl(\mu_*^{(d)},\,\sigma_*^2\bigr),\\
\mu_*^{(d)} &:=\mathbf k_*^\top(\mathbf{K}_{N}+\sigma^2\mathbf I_N)^{-1}\boldsymbol z^{(d)},\\
\sigma_*^2 &:= k_{**}-\mathbf k_*^\top(\mathbf{K}_{N}+\sigma^2\mathbf I_N)^{-1}\mathbf k_*.
\end{aligned}
\end{equation*}
Collecting coordinates, $\mathbf f_*\mid\mathcal D\sim\mathcal N(\boldsymbol\mu_*,\boldsymbol\Sigma_*)$ with $\boldsymbol\mu_*:=(\mu_*^{(1)},\ldots,\mu_*^{(D)})^\top$ and $\boldsymbol\Sigma_*:=\sigma_*^2\mathbf I_D$.

\paragraph{Predictions in probability space}
Given the Gaussian predictive latent distribution $\mathbf f_*\mid\mathcal D\sim\mathcal N(\boldsymbol\mu_*,\boldsymbol\Sigma_*)$ from the previous paragraph, we map samples back to the simplex and estimate the predictive probabilities as an expectation with Monte Carlo,
\begin{equation}
\begin{aligned}
\mathbb E[\boldsymbol\pi_*\mid \boldsymbol{x}_*,\mathcal D] &\approx
\frac{1}{S}\sum_{s=1}^S \varphi^{-1}\!\bigl(\mathbf f_*^{(s)}\bigr) \\
\mathbf f_*^{(s)}&\sim\mathcal N(\boldsymbol\mu_*,\boldsymbol\Sigma_*).
\end{aligned}
\label{eq:gpilr_pred}
\end{equation}
We then predict the class via $\hat c=\arg\max_k \mathbb E[\pi_{*,k}\mid \boldsymbol{x}_*,\mathcal D]$.

\subsection{Sparse Gaussian Processes} \label{sec:sparse}
Exact GP inference scales cubically in $N$ and is prohibitive for large datasets. We therefore use inducing-point approximations; in what follows we outline the two sparse alternatives used in our experiments: an uncollapsed sparse variational classifier (\ourmethodvariational) and a collapsed sparse GP model (\ourmethodsparse). Let $\mathbf X_{\mathrm u}:=\{\boldsymbol{x}^{\mathrm u}_m\}_{m=1}^M$ be inducing inputs and, for each latent GP output $f^{(d)}$, define inducing variables $\mathbf u^{(d)}:=f^{(d)}(\mathbf X_{\mathrm u})$. The inducing locations $\mathbf X_{\mathrm u}$ are treated as parameters and optimized jointly with kernel hyperparameters.

\paragraph{Sparse variational GP Classification (\ourmethodvariational)}
We consider the sparse variational GP classifier of \citet{Hensman2015}, which is often informally referred to as “uncollapsed” because its lower bound is additive over data points, enabling stochastic optimization. 
Instead of using the softmax as the link function we  propose using the bijection $\varphi$. This choice again reduces to only $D$ latent GPs, compared to the standard softmax parameterization (see Eq.~\ref{eq:softmax}). The class probabilities are parameterized as $\boldsymbol\pi(\boldsymbol{x})=\varphi^{-1}(\mathbf f(\boldsymbol{x}))$ for $\mathbf f(\boldsymbol{x})\in\mathbb R^{D}$.
Using inducing variables $\mathbf u:=(\mathbf u^{(1)},\ldots,\mathbf u^{(D)})$, we use a mean-field Gaussian variational posterior $q(\mathbf u)=\prod_{d=1}^D q(\mathbf u^{(d)})$ with $q(\mathbf u^{(d)})=\mathcal N(\mathbf u^{(d)}\mid \mathbf m^{(d)},\mathbf S^{(d)})$. We follow \citet{Hensman2015} to define 
$
q(\mathbf f):=\int p(\mathbf f\mid \mathbf u)\,q(\mathbf u)\,d\mathbf u,
$
which implies Gaussian marginals $q(\mathbf f_i)$ at the training inputs. Parameters $\{\mathbf m^{(d)},\mathbf S^{(d)}\}$, $\mathbf X_{\mathrm u}$, $\theta$, are learned by maximizing the stochastic variational objective
\begin{multline*}
 \log p(\mathbf c) \geq\sum_{i=1}^N \mathbb E_{q(\mathbf f_i)}\bigl[\log p(c_i\mid \mathbf f_i)\bigr]\\
-\sum_{d=1}^D \mathrm{KL}\bigl[q(\mathbf u^{(d)})\,\|\,p(\mathbf u^{(d)})\bigr].
\end{multline*}
We optimize this objective using mini-batches; the expectation under the categorical likelihood $p(c_i\mid \mathbf f_i)=\mathrm{Cat}(\varphi^{-1}(\mathbf f_i))$ is approximated numerically by Monte Carlo. At prediction time, $q(\mathbf f_*)$ is obtained from the variational posterior (a Gaussian with closed-form mean/covariance), and probabilities follow from Monte Carlo through $\varphi^{-1}$. We refer to this model as \ourmethodvariational. 

In our experiments, we also include an additional learnable weight matrix $\mathbf W$  \citep{alvarez2012kernels}. For softmax-based classifiers we use $\boldsymbol\pi(\boldsymbol{x})=\mathrm{softmax}(\mathbf{W}\mathbf f(\boldsymbol{x}))$, and in our case we use $\boldsymbol\pi(\boldsymbol{x})=\varphi^{-1}(\mathbf{W}\mathbf f(\boldsymbol{x}))$. This parameter is optimized jointly with the others w.r.t. the lower bound. 

\paragraph{Sparse GP (\ourmethodsparse)}
For our ILR model, the bijection and our modeling choice give a latent Gaussian likelihood in $\mathbb R^D$: $p(\boldsymbol{z}\mid \mathbf f(\boldsymbol{x}))=\mathcal N(\boldsymbol{z} \mid \mathbf f(\boldsymbol{x}),\sigma^2\mathbf I_D)$. We can therefore follow \citet{milios2018dirichlet} and use the approach of \citet{titsias2009variational} to construct a "collapsed" variational approximation. It uses $D$ latent GPs (one per ILR coordinate) and inducing variables $\boldsymbol{u}$. 
The inducing points together with the model parameters are optimized with the lower bound: 
\begin{multline*}
\log p(\mathbf z)
\;\ge\;
\log \mathcal N\!\left(
\mathbf z \mid \mathbf 0,\,
\mathbf Q_{N}
+ \sigma^2 \mathbf I_N
\right)\\
\;-\;
\frac{1}{2\sigma^2}
\operatorname{tr}\!\left(
\mathbf K_{N} - \mathbf Q_{N}
\right),
\end{multline*}
where $\mathbf Q_{N} = \mathbf K_{NM}\mathbf K_{M}^{-1}\mathbf K_{MN}$.
This reduces the cost to $\mathcal{O}(NM^2)$, we note that we could also have reduced the cost by
introducing minibatches with an additional Gaussian approximation as in \citet{Hensman2013} but decided to use the method of \citet{titsias2009variational} to keep the results comparable to \citet{milios2018dirichlet}.

\section{Related Work}

The standard multi-class GP classifier combines $K$ latent GPs with a softmax link \citep{williams1998bayesian}, which yields a non-conjugate likelihood and motivates approximate inference. Classic approximations include Laplace and related methods \citep{nickisch2008approximations}, as well as expectation propagation \citep{hernandez2016scalable, villacampa2017scalable} and variational inference with inducing points \citep{Hensman2015, hensman2015mcmc}. 
Recent work has continued to expand the modeling and scalability of multi-class GP classification, for example by handling various likelihoods \citep{Liu2019}, incorporating input noise \citep{villacampa2021multi}, or using transformed constructions for non-stationary and dependent outputs \citep{maronas2023efficient}. There is also an active line of work on deep GP models for classification \citep{blomqvist2019deep, dutordoir2020bayesian} and on robustness considerations \citep{hernandez2011robust, blaas2020adversarial}. There are GP classification models that achieve conditional conjugacy by data-augmentation \citep{wenzel2019efficient, Galy2020, achituve2021gp}. \citet[Chapter~7]{galyfajou2022latent} realized that the method of \citet{Galy2020} can use $D$ latent GPs instead of $K$. They consider a different link function from ours,  we use the ILR bijection and bypass the need for auxiliary variables and Gibbs sampling, making the classification problem a regression problem in the latent space.

\section{Experiments}
\begin{figure}
    \centering
    \includegraphics[width=0.32\linewidth]{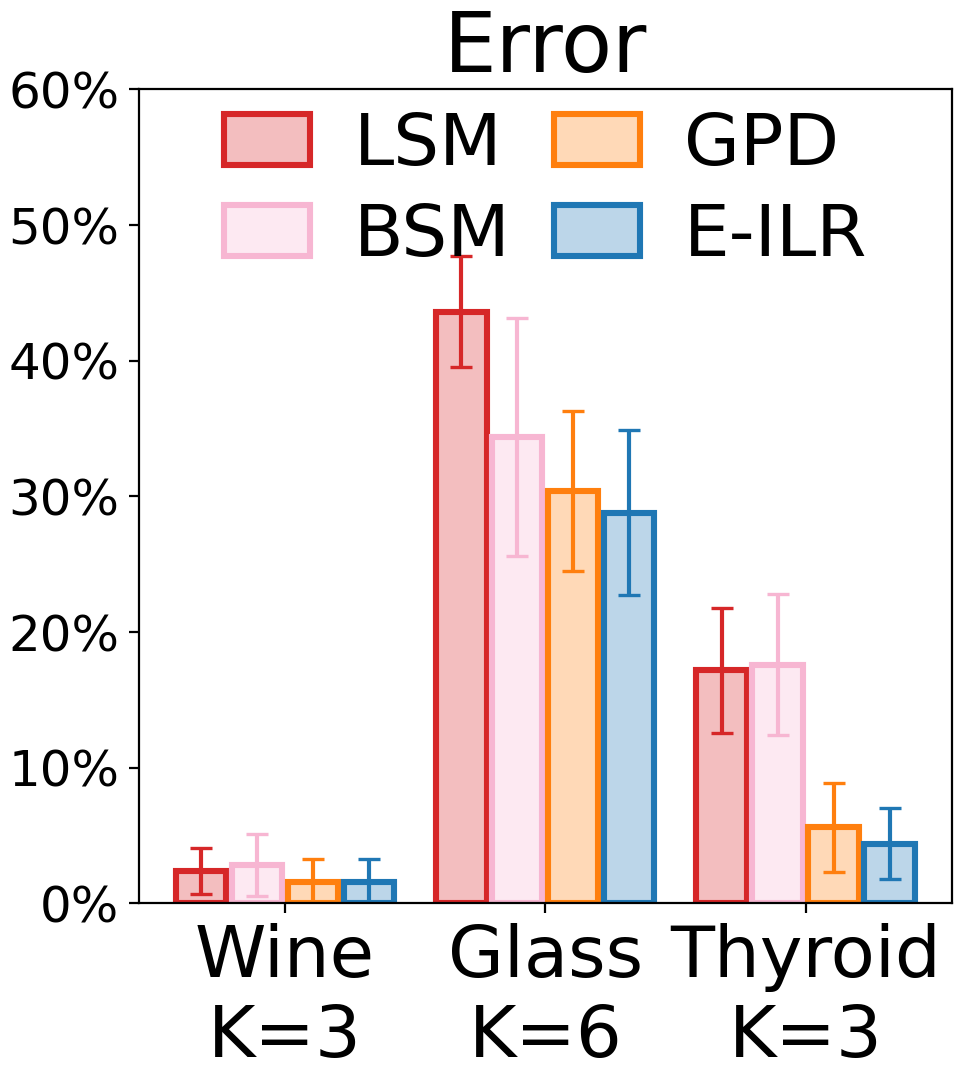}
    \includegraphics[width=0.32\linewidth]{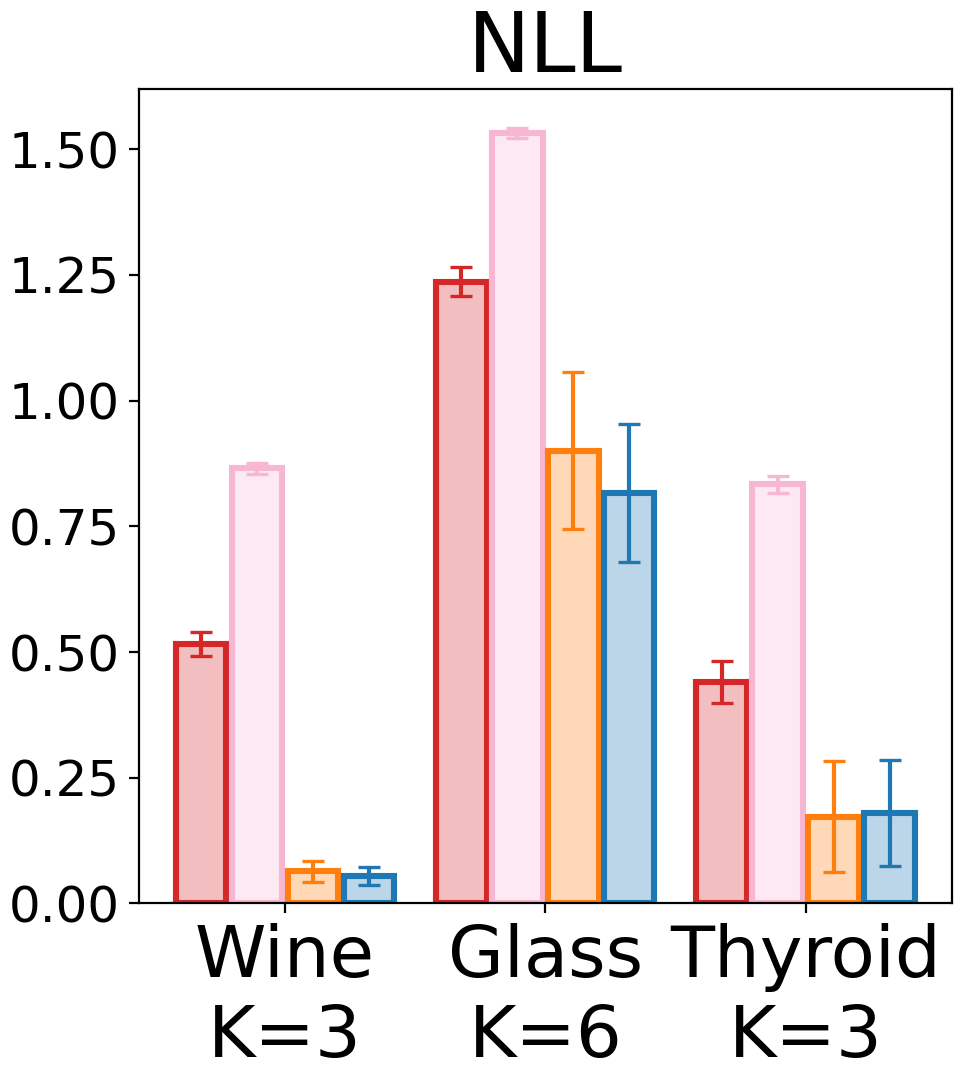}
    \includegraphics[width=0.32\linewidth]{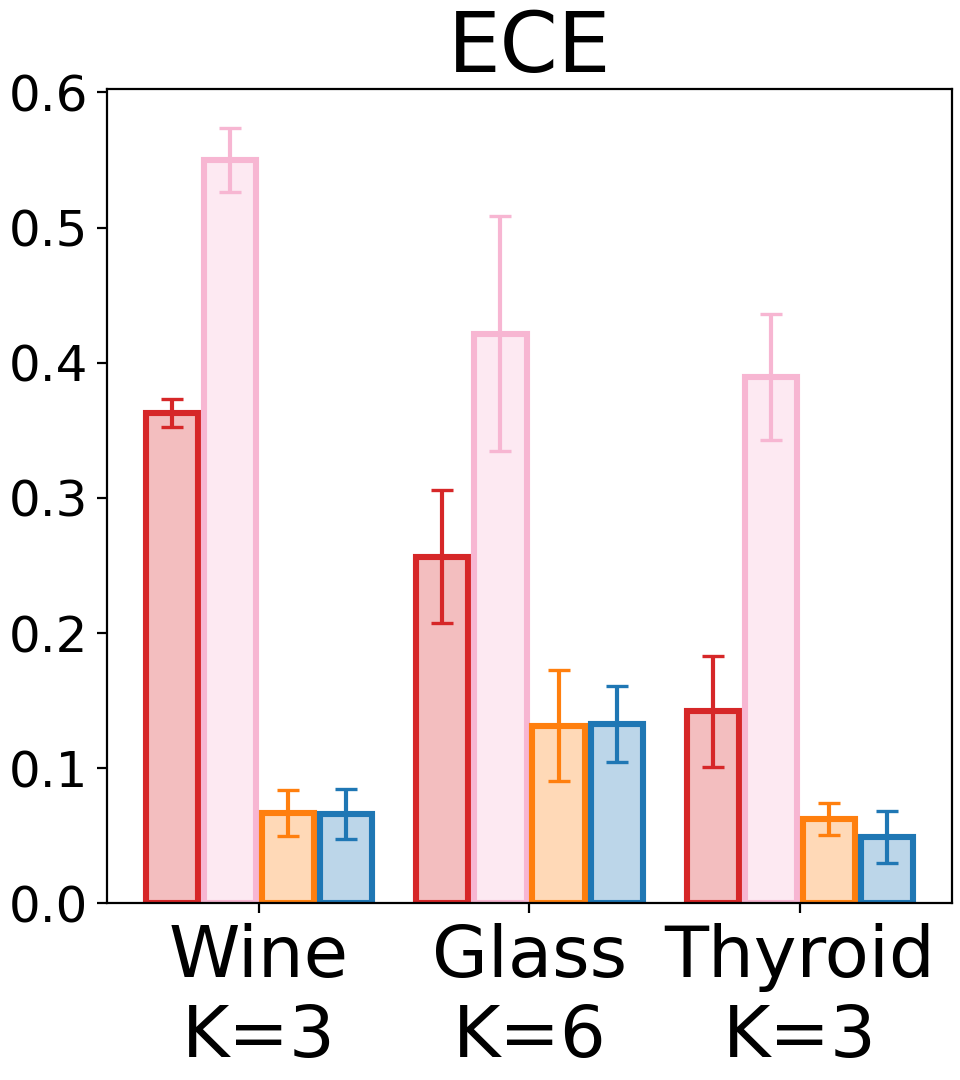}
    \caption{Error, NLL and ECE for UCI datasets in the exact setting.  }
    \label{fig:uci_conjugate}
\end{figure}
\begin{figure*}
    \centering
    \begin{subfigure}[b]{0.33\linewidth}
        \centering
        \includegraphics[width=\linewidth]{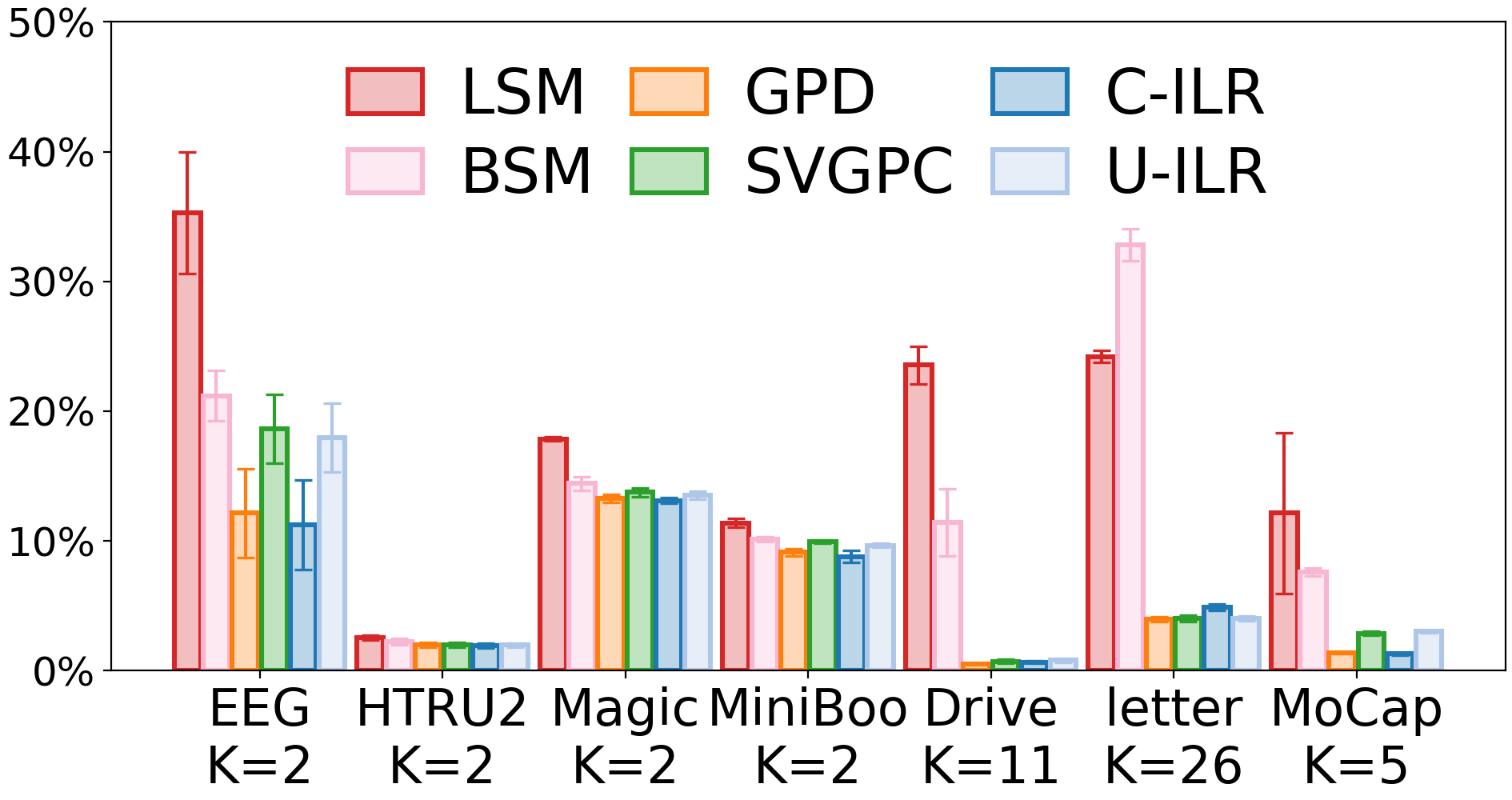}
        \caption{Error rate}
    \end{subfigure}
    \begin{subfigure}[b]{0.33\linewidth}
        \centering
        \includegraphics[width=\linewidth]{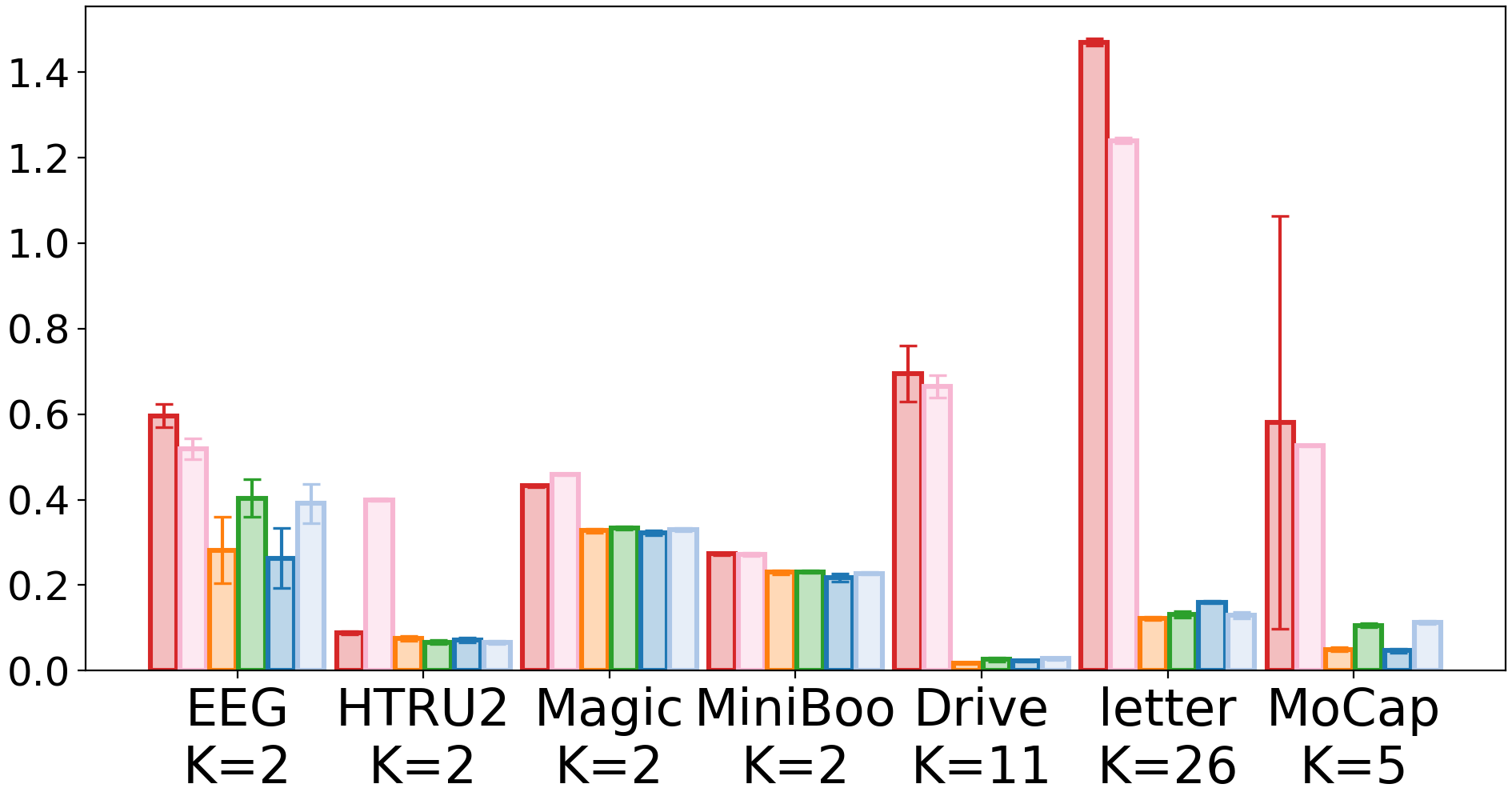}
        \caption{Negative log-likelihood}
    \end{subfigure}
    \begin{subfigure}[b]{0.33\linewidth}
        \centering
        \includegraphics[width=\linewidth]{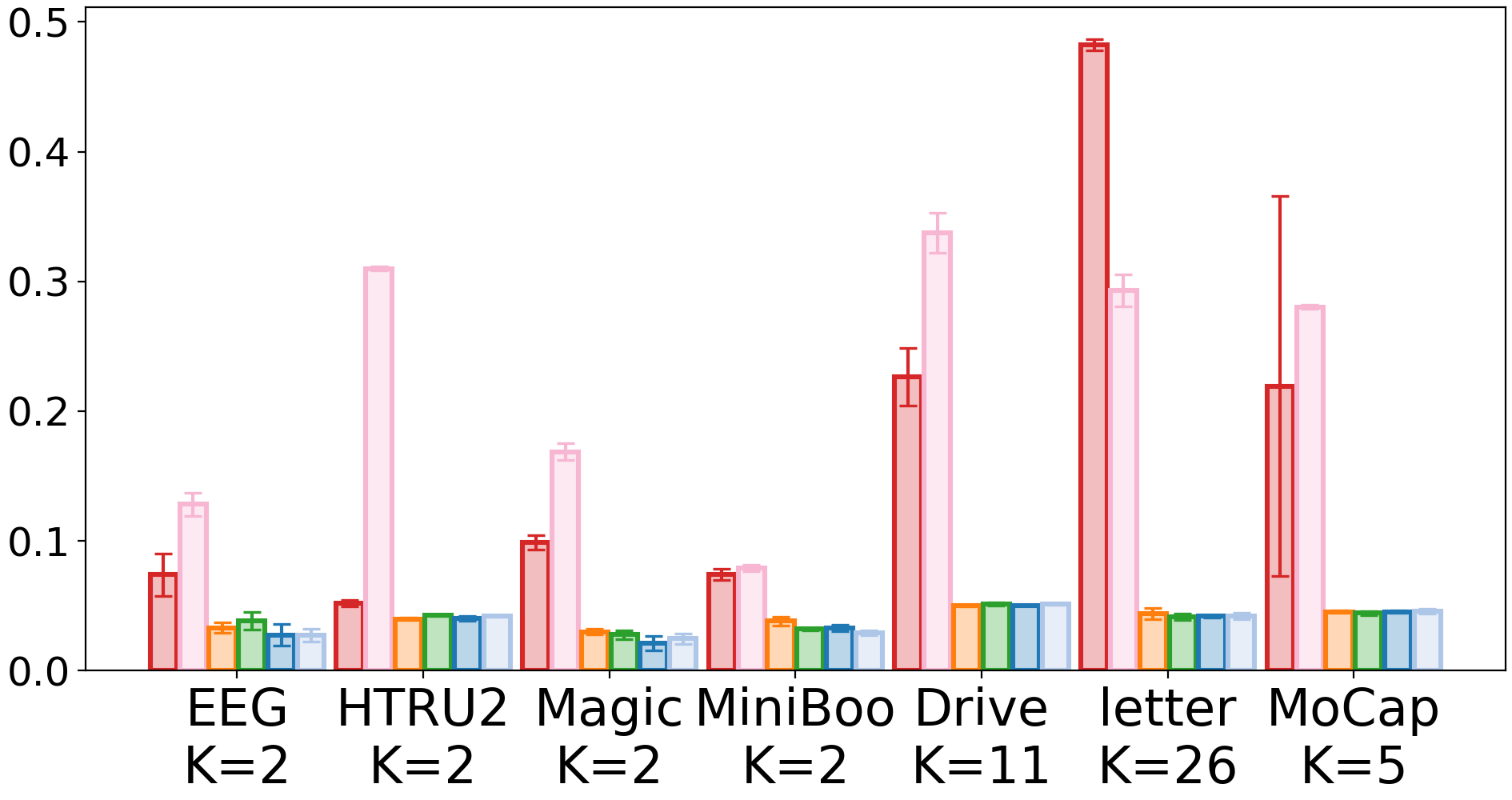}
        \caption{Expected Calibration Error}
    \end{subfigure}    
    \caption{Sparse models performance on datasets from the UCI repository.  }
    \label{fig:uci_sparse}
\end{figure*}

We evaluate our approach on synthetic and real-world benchmarks to assess predictive accuracy and calibration.  We compare against Dirichlet-based Gaussian processes (GPD) \citep{milios2018dirichlet}, Logistic Softmax (LSM) \citep{Galy2020}, Bijective Softmax (BSM) \citep{galyfajou2022latent}, and sparse variational Gaussian process classification (SVGPC) \citep{Hensman2015}. 
Throughout the experiments, ``exact'' denotes the non-sparse setting, where posterior inference is done directly for the corresponding latent/augmented model: for our method and GPD via direct Gaussian latent posterior computations, and for LSM/BSM via Gibbs sampling in the augmented model. ``sparse'' denotes inducing-point approximations.  For GPD, LSM and BSM we consider both exact and sparse settings.
For our method, we evaluate one exact variant (\ourmethod) and two sparse variants (\ourmethodsparse\ and \ourmethodvariational). In the plots, for space reasons, we use the short labels E-ILR, C-ILR, and U-ILR for \ourmethod, \ourmethodsparse and \ourmethodvariational. The implementation code is available at \href{https://github.com/williwilliams3/gpilr}{github.com/williwilliams3/gpilr}..

We use independent zero-mean GP priors with an RBF kernel $K_{\theta}$ for all methods,
$
K_{\theta}(x,x')=\sigma_f^2\exp\!\left(-\frac{\|x-x'\|_2^2}{2\ell^2}\right),
$
with hyperparameters $\theta=(\sigma_f^2,\ell)$.

We report accuracy, negative log-likelihood (NLL), and expected calibration error (ECE) on a test set. The NLL is the categorical cross entropy $\mathrm{NLL}:=-\frac{1}{n}\sum_{i=1}^n \log \hat\pi_{i,c_i}$. ECE \citep{guo2017calibration} is computed with $M=10$ equally spaced bins over a confidence score. For binary classification we compute the confidence for class $c=1$, and for multi-class classification ($K>2$) we use the standard \emph{top-label} confidence $\hat p_i:=\max_k \hat\pi_{i,k}$, yielding
\begin{equation*}
    \mathrm{ECE}:=\sum_{m=1}^{M} \frac{|B_m|}{n}\,\big|\mathrm{acc}(B_m)-\mathrm{conf}(B_m)\big|,
\end{equation*}
where $\mathrm{conf}(B_m):=\frac{1}{|B_m|}\sum_{i\in B_m}\hat p_i$.
We fix the tolerance parameter $\varepsilon=10^{-6}$ (as in Proposition~\ref{prop:sigma_bound}) for all experiments. Unless stated otherwise, we report results as the mean and one standard deviation over 5 random train/validation/test splits (seeds).

\paragraph{UCI classification in the exact setting}
We evaluate methods in the exact setting on three small UCI datasets from \citet{hernandez2011robust}, each with fewer than 300 samples, where exact $\mathcal O(N^3)$ inference is feasible. For each seed, we set aside 50 points as a test set and use 10\% of the remaining training data as a validation set. We tune the label-smoothing parameter over $\lambda\in\{0.95,0.99,0.999,0.9999\}$ and select the best value based on validation NLL. For GPD, we tune $\alpha_{\varepsilon}\in\{0.1,0.01,0.001,0.0001\}$ and select the model's parameters by $\tfrac{1}{2}(\mathrm{NLL}_{\mathrm{val}} + \mathrm{ECE}_{\mathrm{val}})$.

Figure~\ref{fig:uci_conjugate} shows that our method and GPD achieve similar accuracy, while our method attains slightly lower NLL on two of the three datasets and the lowest ECE on all three, meaning it is slightly better calibrated. LSM and BLM perform worse than the non-augmented models.

\paragraph{UCI classification benchmarks in the sparse setting}
We replicate the UCI benchmarks from \citet{milios2018dirichlet} on 7 classification tasks, which require sparse inference due to dataset sizes ranging from tens of thousands to hundreds of thousands of points. We use a validation set consisting of 10\% of the training data to select hyperparameters. We tune $\lambda\in\{0.95,0.99,0.999,0.9999,0.999999\}$ and choose the value that minimizes validation NLL; for GPD we tune $\alpha_{\varepsilon}\in\{0.1,0.01,0.001,0.0001\}$ and select by the validation loss  $\tfrac{1}{2}(\mathrm{NLL}_{\mathrm{val}} + \mathrm{ECE}_{\mathrm{val}})$. For optimization, we try learning rates $10^{-2}$ and $10^{-3}$ for all methods, and we report the best setting on the validation set. We additionally consider two standard input preprocessing  variants ($-1$ \emph{to} $1$) and \emph{unit-variance} normalization) for all methods.

We initialize inducing points with $k$-means++ \citep{arthur2006k}; the number of inducing points and dataset-specific details follow \citet{milios2018dirichlet} and are reported in the appendix. For \compmethodb\ and \ourmethodvariational, we include an additional learnable mixing matrix $\boldsymbol W$ before the link function (Sec.~\ref{sec:sparse}). For sparse "collapsed" GPD and sparse \ourmethodsparse, we also consider a more flexible likelihood by adding an extra per-class scale parameter to the observation noise.

Figure~\ref{fig:uci_sparse} summarizes the results (see App.~\ref{app:exp} for the detailed numerical values). Overall, sparse GPD and \ourmethodsparse\ achieve strong performance across metrics. The ILR-based sparse variants (\ourmethodsparse\ and \ourmethodvariational) are better in terms of ECE on 3 out of the 7 tasks and are otherwise comparable to their softmax/Dirichlet counterparts. On the \textsc{Letter} dataset \ourmethodvariational\ is better than \ourmethod.
The data-augmented methods LSM and BSM struggle across all metrics, especially for more than 2 categories. In summary, our sparse methods are similar in performance to sparse GPD, while sometimes being slightly better calibrated.

\paragraph{Effect of the parameter $\lambda$}
\begin{figure}
    \centering
    \includegraphics[width=0.32\linewidth]{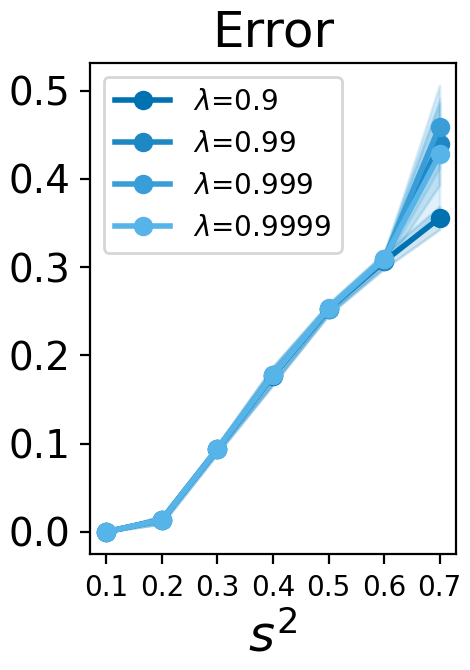}
    \includegraphics[width=0.32\linewidth]{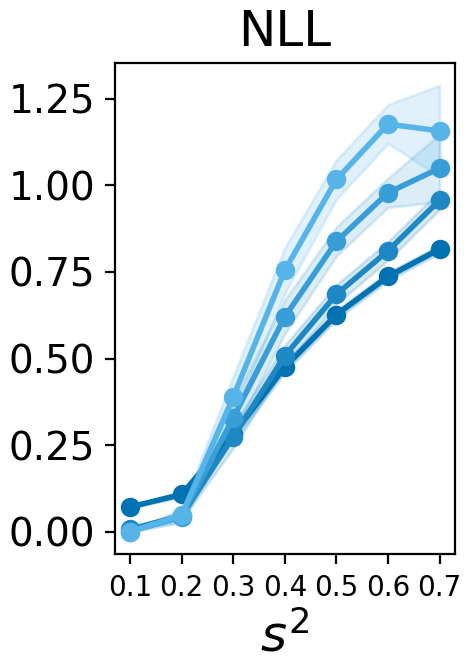}
    \includegraphics[width=0.32\linewidth]{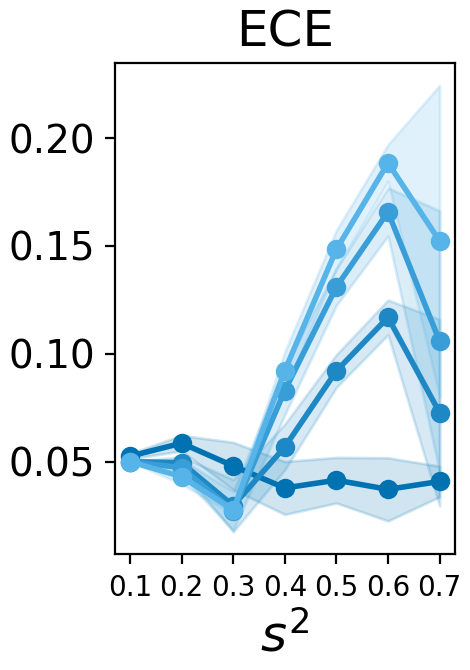}
    \caption{Error, NLL and ECE for increasing overlap of the input variables. }
    \label{fig:toy}
\end{figure}
We replicate the setup of \citet{Galy2020} to evaluate how the label-smoothing parameter $\lambda$ affects calibration as the class assignment becomes less clear. We consider a synthetic dataset with $K=3$ classes where the covariates are generated from a mixture model (one component per class) and control class ambiguity by varying the component standard deviation $s\in\{0.1,\ldots,0.7\}$. As $s^2$ increases, the components overlap and the conditional class probabilities become close to uniform over large parts of the covariate space (e.g., for $s=0.7$ almost all inputs are plausibly generated by multiple components). In this regime, a well-calibrated classifier should output near-uniform predictive probabilities except in small regions where one component clearly dominates and the model should be confident.

Fig.~\ref{fig:toy} shows that NLL and ECE vary with $\lambda$, especially at higher overlap. Across noise levels $s^2$, we find that $\lambda=0.9$ yields the most stable calibration and best error and NLL, whereas larger values tend to become overconfident when the inputs are ambiguous. For lower noise levels (small $s^2$), higher values of $\lambda$ perform better, as the model can be more confident. Since the optimal $\lambda$ is problem-dependent, in the remaining experiments we select $\lambda$ using a validation set.

\paragraph{Scaling with the number of categories}
\begin{figure}
    \centering
    \includegraphics[width=0.32\linewidth]{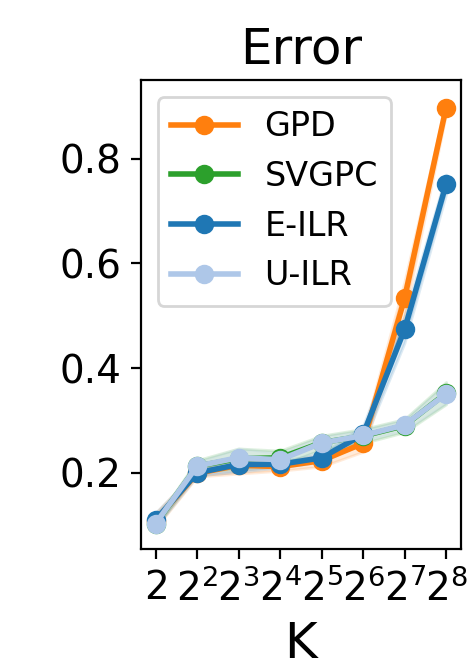}
    \includegraphics[width=0.32\linewidth]{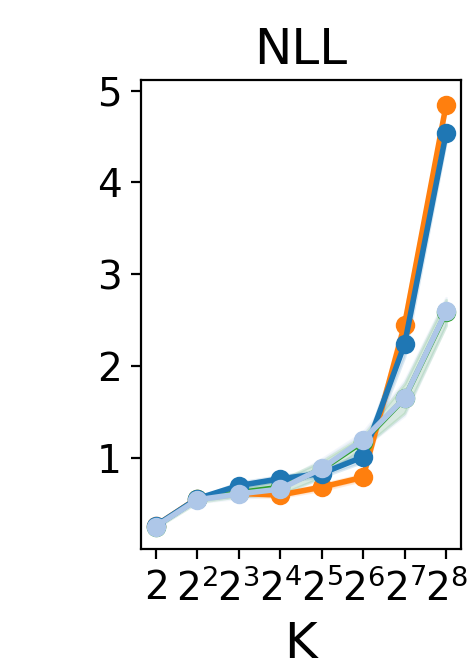}
    \includegraphics[width=0.32\linewidth]{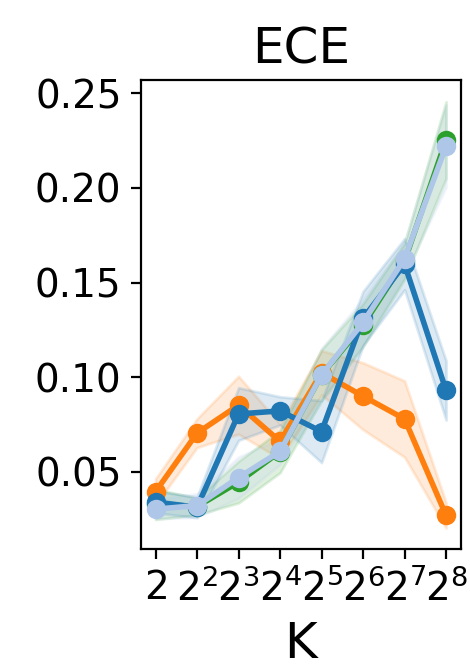}
    \caption{Error, NLL and ECE for increasing number of categories $K$. }
    \label{fig:scaling}
\end{figure}
We next study how performance changes as the number of classes grows. For $K\in\{2,4,\ldots,256\}$, we generate a $2$-dimensional input classification problem where each class corresponds to one component in a Gaussian mixture with means placed evenly around the unit circle. We choose a shared variance so that nearby components have non-trivial overlap; this ensures that the Bayes-optimal classifier is not always certain and that a well-calibrated model should express uncertainty in regions where multiple classes are plausible.

For each value of $K$, we use $N=1000$ training points and $1000$ test points. We tune the label-smoothing parameter over $\lambda\in\{0.95,0.99,0.999,0.9999\}$ and report the value that minimizes training NLL. For GPD, we similarly tune the concentration parameter over $\alpha_{\varepsilon}\in\{0.1,0.01,0.001,0.0001\}$ using training NLL. For the sparse variational baselines (\compmethodb\ and \ourmethodvariational), we use $M=128$ inducing points selected with $k$-means++.

Figure~\ref{fig:scaling} summarizes the results. \compmethodb\ and \ourmethodvariational\ exhibit very similar performance across $K$, indicating that the ILR parameterization works as well as the sparse variational baseline, and they are the most accurate for $K>2^6$.
\ourmethod\ is as accurate as GPD for $K\leq 2^6$ and slightly more accurate for $K>2^6$. It is better calibrated than GPD for $K=2^2$ to $2^5$ but less calibrated for $K>2^6$.

\paragraph{When does GPD break down?}
This experiment highlights an often-overlooked choice in Dirichlet-based GP classification (GPD): whether to form predictive probabilities from the latent GP predictive $p(\mathbf{f}_*\mid \mathcal D,\boldsymbol{x}_*)$ or from the predictive distribution of the noisy pseudo-observation $p(\boldsymbol{z}_*\mid \mathcal D, \boldsymbol{x}_*)$ (i.e., including the likelihood noise). Concretely, in GPD one can estimate probabilities either as $\boldsymbol\pi_*\approx \mathrm{softmax}(\mathbf{f}_*)$ with $\mathbf{f}_*\sim p(\mathbf{f}_*\mid\mathcal D, \boldsymbol{x}_*)$, or as $\boldsymbol\pi_*\approx \mathrm{softmax}(\boldsymbol{z}_*)$ with $\boldsymbol{z}_*\sim p(\boldsymbol{z}_*\mid\mathcal D,\boldsymbol{x}_*)$. For our method, the analogous choice is whether to apply $\varphi^{-1}$ to samples from $p(\mathbf{f}_*\mid\mathcal D,\boldsymbol{x}_*)$ or from $p(\boldsymbol{z}_*\mid\mathcal D,\boldsymbol{x}_*)$.

We consider the same synthetic setup as the \citet{Galy2020} and focus on an easy regime where both models fit perfectly when using the latent predictive: we fix $\lambda=0.9$, set $\alpha_{\varepsilon}=0.01$, and choose $s=0.1$ so the mixture components are well separated. Table~\ref{tab:gpd_noise_pred} reports test error when we estimate $\boldsymbol\pi_*$ using a single Monte Carlo sample (Eq.~\ref{eq:gpilr_pred}) under the two choices. While both methods achieve zero error when predicting via $p(\mathbf f_*\mid\mathcal D,\boldsymbol{x}_*)$, using $p(\boldsymbol{z}_*\mid\mathcal D, \boldsymbol{x}_*)$ can introduce avoidable misclassifications for GPD.

This behavior is a consequence of estimating the probabilities from noisy pseudo-observations under the log-normal approximation: sampling $\boldsymbol{z}_*$ adds variability before the softmax, and fluctuations can move a sample into a different $\arg\max$ region. The size of this effect depends on both $K$ and $\alpha_{\varepsilon}$. As $K$ increases, there are more competing classes, so the probability that \emph{some} incorrect coordinate becomes the maximum increases. Meanwhile, our numerical estimates (see Appendix~\ref{app:exp}) show that the error from using $p(\boldsymbol{z}_*\mid\mathcal D,\boldsymbol{x}_*)$ decreases as $\alpha_{\varepsilon}$ becomes smaller. In practice, averaging over multiple samples reduces the effect, and using the latent predictive $p(\mathbf{f}_*\mid\mathcal D,\boldsymbol{x}_*)$ works well for both methods. For \ourmethod, we additionally choose $\sigma^2$ to make the probability of crossing decision boundaries in latent space negligible (Proposition~\ref{prop:sigma_bound}), which explains the robustness observed here.

\begin{table}[t]
\centering
\begin{tabular}{lcc}
\toprule
Model &   Error ($y_*$) &  Error ($f_*$) \\
\midrule
GPD & $0.076 \pm 0.013$ & $0.0 \pm 0.0$ \\
\ourmethod & $0.0 \pm 0.0$ & $0.0 \pm 0.0$ \\
\bottomrule
\end{tabular}
\caption{Test misclassification rate when forming $\boldsymbol\pi_*$ from either the noisy predictive distribution $p(\boldsymbol{z}_*\mid\mathcal D,\boldsymbol{x}_*)$ or the latent predictive $p(\mathbf{f}_*\mid\mathcal D,\boldsymbol{x}_*)$. For GPD we compute $\boldsymbol\pi_* = \mathrm{softmax}(\cdot)$, applying softmax to samples of either $\boldsymbol z_*$ or $\mathbf f_*$. For \ourmethod\ we compute $\boldsymbol\pi_* = \varphi^{-1}(\cdot)$, applying $\varphi^{-1}$ to samples of either $\boldsymbol z_*$ or $\mathbf f_*$.}
\label{tab:gpd_noise_pred}
\end{table}

\section{Conclusion}

We introduced a new conjugate multiclass Gaussian process classification model that represents class probabilities on the simplex and uses the ILR bijection to work in a $D=K-1$ dimensional Euclidean latent space. This turns classification into standard multi-output GP regression with a Gaussian likelihood on latent targets, yielding closed-form posterior inference and exact marginal-likelihood learning with $\mathcal O(N^3)$ complexity in the number of training points. 
Importantly, our results highlight a clear empirical gap between simplex-based models (\ourmethod, \ourmethodsparse, and GPD) and auxiliary-variable and sparse variational GP classifiers in both the exact and sparse settings. Across experiments, the simplex-based models consistently achieve stronger accuracy, calibration, and robustness, while auxiliary-variable methods generally lag behind and sparse variational approaches can underperform in some scenarios.
This motivates the use of our method on tasks where reliable uncertainty calibrated estimates are crucial. Within the class of well-calibrated, conjugate models, our method offers a principled alternative to Dirichlet-based GP classification, achieving conjugacy without relying on the log-normal approximation to the Gamma construction, while delivering comparable or slightly improved calibration without sacrificing accuracy.

\paragraph{Discussion and outlook.}
From a methodological standpoint, reducing multiclass classification to GP regression in latent ILR coordinates provides a simple and scalable framework. Because the core problem becomes standard multi-output GP regression, existing advances in scalable GP inference can be used directly. In this work we employed inducing-point variational methods \citep{titsias2009variational, Hensman2015}, but the same formulation is compatible with more recent sparse and variational constructions \citep{salimbeni2018orthogonally, hensman2018variational, shi2020sparse, rossi2021sparse, wenger2022posterior}, iterative linear-algebra solvers for approximate GP inference \citep{wang2019exact, wilson2021pathwise, wenger2022preconditioning, lin2024stochastic}, and structured kernel interpolation and related kernel representations \citep{wilson2015kernel, gardner2018product}. Our formulation  introduces only a small number of interpretable design choices, notably the label-smoothing parameter $\lambda$ and the latent noise level $\sigma^2$, which we selected using validation or principled overlap bounds; future work could consider adaptive (e.g., input or class dependent) choices while retaining tractable inference. Overall, the proposed construction provides a modular way to combine well-calibrated multiclass GP classification with  GP regression tools, including scalable solvers, structured kernels, and richer priors, in larger and more complex settings.

\begin{acknowledgements} 
The authors were supported by the Research Council of Finland Flagship programme: Finnish Center for Artificial Intelligence FCAI, and additionally by grants: 363317 (BW, AK), 369502 (MH).
The authors acknowledge the research environment provided by ELLIS Institute Finland, and CSC - IT Center for Science, Finland for computational resources.
\end{acknowledgements}

\bibliography{bibtex}

\newpage

\onecolumn

\title{
Simplex-to-Euclidean Bijection for Conjugate and Calibrated Multiclass Gaussian Process Classification
\\(Supplementary Material)}
\maketitle
\appendix

\section{Mathematical Derivations and Training Algorithms}
\subsection{Proof of Proposition~\ref{prop:sigma_bound}}\label{app:proof_sigma_bound}
The proof follows three steps: (a) computing the pairwise separation of the centers, (b) bounding the misassignment probability for a single competitor center and (c) apply the union bound for all competitors for computing the upper bound on $
\sigma$. 

\paragraph{Pairwise separation of the centers.}
Write
\[
a:=\lambda+\frac{1-\lambda}{K},\qquad b:=\frac{1-\lambda}{K},
\qquad L:=\log\frac{a}{b}=\log\!\left(1+\frac{K\lambda}{1-\lambda}\right).
\]
For $k\neq \ell$, the compositions $\boldsymbol\mu^{(k)}$ and $\boldsymbol\mu^{(\ell)}$ differ only in the $k$-th and $\ell$-th coordinates.
We will compute the distance between them as 
$\norm{\boldsymbol\mu^{(k)} - \boldsymbol\mu^{(\ell)}}^2_A = \norm{\boldsymbol\mu^{(k)}}^2_A + \norm{\boldsymbol\mu^{(\ell)}}^2_A - 2\langle \boldsymbol\mu^{(k)}, \boldsymbol\mu^{(\ell)} \rangle_A $

\begin{equation*}
\langle \boldsymbol\mu^{(k)}, \boldsymbol\mu^{(\ell)} \rangle_A := \frac{1}{2K} \sum_{i,j=1}^{K} 
     \log\frac{ \mu^{(k)}_i}{ \mu^{(k)}_j}\log\frac{ \mu^{(\ell)}_i}{ \mu^{(\ell)}_j} 
     = \frac{1}{K}\log \frac{a}{b}\log \frac{b}{a} 
     = -\frac{1}{K}L^2
\end{equation*}
Appendix~A of \citet{Williams2026} showed that $\norm{\boldsymbol\mu^{(k)}}_A^2 = \frac{D}{K}L^2$, then putting it all together we obtain
\begin{equation*}
    \norm{\boldsymbol\mu^{(k)} - \boldsymbol\mu^{(\ell)}}^2_A = 2 \frac{D}{K}L^2 + \frac{2}{K}L^2
     = 2 L^2.
\end{equation*}

Since the ILR transform $\varphi$ is an isometry between the Aitchison geometry on $\mathring\Delta^{D}$ and the Euclidean
geometry on $\mathbb R^{D}$,
 then $\|\boldsymbol\mu^{(k)}-\boldsymbol\mu^{(\ell)}\|_A=\|\varphi(\boldsymbol\mu^{(k)})-\varphi(\boldsymbol\mu^{(\ell)})\|_2$, the Euclidean distance between the ILR images equals the above Aitchison distance.
Define $\boldsymbol m^{(k)} = \varphi(\boldsymbol\mu^{(k)})$, 
\begin{equation}\label{eq:delta_def}
\delta \;:=\; \|\boldsymbol m^{(k)}-\boldsymbol m^{(\ell)}\| \;=\; \sqrt{2}\,L
\;=\;\sqrt{2}\,\log\!\left(1+\frac{K\lambda}{1-\lambda}\right),\qquad k\neq \ell.
\end{equation}
\paragraph{Misassignment probability for one competitor.}
Fix $k\neq \ell$ and define the unit direction $\boldsymbol u_{k\ell}:=(\boldsymbol m^{(\ell)}-\boldsymbol m^{(k)})/\|\boldsymbol m^{(\ell)}-\boldsymbol m^{(k)}\|$.
Thus, for $\boldsymbol Z\sim\mathcal N(\boldsymbol m^{(k)},\sigma^2 \mathbf I_D)$,
\[
\{\boldsymbol Z \text{ is closer to } \boldsymbol m^{(\ell)} \text{ than to } \boldsymbol m^{(k)}\}
\; = \;
\Big\{\boldsymbol u_{k\ell}^\top(\boldsymbol Z-\boldsymbol m^{(k)})\ge \frac{\delta}{2}\Big\}.
\]
But $\boldsymbol u_{k\ell}^\top(\boldsymbol Z-\boldsymbol m^{(k)})\sim\mathcal N(0,\sigma^2)$, hence
\begin{equation}\label{eq:pair_error}
\mathbb P\!\big(\|\boldsymbol Z-\boldsymbol m^{(\ell)}\|\le \|\boldsymbol Z-\boldsymbol m^{(k)}\| \,\big|\, C=k,\boldsymbol{x}\big)
\;\le\;
\Phi\!\left(-\frac{\delta}{2\sigma}\right).
\end{equation}
\paragraph{Union bound over $D$ competitors and choice of $\sigma$.}
Let $\mathcal V_k:=\{\boldsymbol z\in\mathbb R^{D}:\|\boldsymbol z-\boldsymbol m^{(k)}\|\le \|\boldsymbol z-\boldsymbol m^{(\ell)}\|\ \forall \ell\}$.
The event $\{\boldsymbol Z\notin\mathcal V_k\}$ implies that there exists some $\ell\neq k$ such that
$\|\boldsymbol Z-\boldsymbol m^{(\ell)}\|\le \|\boldsymbol Z-\boldsymbol m^{(k)}\|$. Therefore, using \eqref{eq:pair_error} and the union bound,
\[
\mathbb P(\boldsymbol Z\notin\mathcal V_k\mid C=k,\boldsymbol{x})
\;\le\;
\sum_{\ell\neq k} \Phi\!\left(-\frac{\delta}{2\sigma}\right)
\;=\;
D\,\Phi\!\left(-\frac{\delta}{2\sigma}\right).
\]
It suffices to impose $D\,\Phi(-\delta/(2\sigma))\le \varepsilon$, i.e.
\[
\Phi\!\left(\frac{\delta}{2\sigma}\right)\ge 1-\frac{\varepsilon}{D}
\quad\Longleftrightarrow\quad
\frac{\delta}{2\sigma}\ge z_{1-\varepsilon/D}.
\]
Substituting \eqref{eq:delta_def} yields exactly the bound 
\begin{equation}
\sigma \;\le\; \frac{\log\!\left(1+\frac{K\lambda}{1-\lambda}\right)}{\sqrt{2}\; z_{1-\varepsilon/D}},
\qquad z_{q}:=\Phi^{-1}(q),
\end{equation}
completing the proof.
\qed

\subsection{The Helmert matrix}

The Helmert matrix $\mathbf{H}\in\mathbb{R}^{D\times K}$ is constructed in \citep{Lancaster1965}. The entries of $\mathbf{H}$ can be written as
$$
H_{ij} =
\begin{cases}
\frac{1}{\sqrt{i(i+1)}}, & j \le i, \\
-\frac{i}{\sqrt{i(i+1)}}, & j = i+1, \\
0, & j > i+1,
\end{cases}
\qquad i=1,\ldots,D.
$$
The rows of $\mathbf{H}$ form an orthonormal basis for the subspace
\begin{equation*}
\mathbb{H}^{D}=\left\{\boldsymbol{z}\in\mathbb{R}^K:\sum_{i=1}^K z_i=0\right\},
\end{equation*}
and therefore satisfy
$
\mathbf{H}^\top \mathbf{H} = \mathbf{I}_{D},
$ and $
\mathbf{H} \mathbf{1} = \mathbf{0}.
$
These properties allow the Helmert matrix to fit naturally inside the isometric logratio, and stablish the isometry between Euclidean coordinates and the simplex coordinates \citep{Egozcue2003}. 

\subsection{Training Algorithms}
Algorithms~\ref{alg:ilr} and~\ref{alg:gpd} summarize the exact conjugate training procedure for our ILR model and for GPD. In both cases the observed class labels are first converted into continuous pseudo-observations, and the Gaussian process marginal likelihood is evaluated on these pseudo-observations, not on the original discrete labels directly.

For \ourmethod, a label $c_i=k$ is replaced by the deterministic ILR target $\boldsymbol z_i=\varphi(\boldsymbol\mu^{(k)})\in\mathbb R^D$, where $\boldsymbol\mu^{(k)}$ is a smoothed one-hot vector. The exact GP marginal likelihood therefore receives the matrix $\mathbf Z\in\mathbb R^{N\times D}$ together with a homoscedastic Gaussian likelihood variance $\sigma^2$. For GPD, the label $c_i=k$ defines Dirichlet parameters $\boldsymbol\alpha_i$. Moment matching the corresponding Gamma variables with log-normal variables gives a pseudo-observation $\tilde{\boldsymbol y}_i\in\mathbb R^K$ and label-dependent variances $\tilde{\boldsymbol\sigma}_i^2\in\mathbb R^K$. Thus GPD maximizes a Gaussian marginal likelihood for $\tilde{\mathbf Y}\in\mathbb R^{N\times K}$ with heteroscedastic diagonal noise. The independent latent GPs share kernel hyperparameters in both algorithms. The smoothing parameters $\lambda$ and $\alpha_\varepsilon$ control how close the pseudo-observations are to the simplex vertices and are chosen by validation in our experiments.

\noindent
\begin{minipage}[t]{0.48\textwidth}
\hrule
\vspace{0.35em}
{\captionsetup{hypcap=false,skip=0pt}\captionof{algorithm}{Exact GP ILR (\ourmethod)}}
\label{alg:ilr}
\vspace{0.35em}
\hrule
\vspace{0.45em}
\normalsize
\begin{algorithmic}[1]
\Require $\mathcal D=\{(\boldsymbol{x}_i,c_i)\}_{i=1}^N$, $K$, $\lambda$, tolerance $\varepsilon$, kernel $K_\theta$
\State $D\gets K-1$; build Helmert matrix $\mathbf H\in\mathbb R^{D\times K}$
\State Choose $\sigma^2$ using Prop.~\ref{prop:sigma_bound} with $\varepsilon$
\For{$i=1,\ldots,N$}
    \State $\boldsymbol\mu^{(c_i)} \gets \lambda\mathbf e_{c_i}+(1-\lambda)\boldsymbol 1/K$ \Comment{Interpolate}
    \State $\boldsymbol z_i \gets \mathbf{H}\log \boldsymbol{\mu}^{(c_i)}$ \Comment{to Euclidean space}    
\EndFor
\For{$d=1,\ldots,D$}
    \State $\boldsymbol z^{(d)}\gets (z_{1d},\ldots,z_{Nd})^\top$
    \State $f^{(d)}(\cdot)\sim\mathcal{GP}(0,K_\theta)$    
\EndFor
\State Optimize shared $\theta$ by maximizing
\Statex $\sum_{d=1}^{D}\log\mathcal N\!\left(\boldsymbol z^{(d)}\mid \mathbf 0,\mathbf K_{\theta}(\mathbf X,\mathbf X)+\sigma^2\mathbf I_N\right)$
\State Predict by sampling $\mathbf f_*$ from the GP posterior and averaging $\varphi^{-1}(\mathbf f_*)$
\end{algorithmic}
\vspace{0.35em}
\hrule
\end{minipage}
\hfill
\begin{minipage}[t]{0.48\textwidth}
\hrule
\vspace{0.35em}
{\captionsetup{hypcap=false,skip=0pt}\captionof{algorithm}{Dirichlet GP (GPD)}}
\label{alg:gpd}
\vspace{0.35em}
\hrule
\vspace{0.45em}
\normalsize
\begin{algorithmic}[1]
\Require $\mathcal D=\{(\boldsymbol{x}_i,c_i)\}_{i=1}^N$, $K$, $\alpha_\varepsilon$, kernel $K_\theta$
\For{$i=1,\ldots,N$}
    \State $\boldsymbol\alpha_{i} \gets \mathbf e_{c_i} + \alpha_\varepsilon \boldsymbol 1 $ 
    \State $ \tilde{\boldsymbol\sigma}^{2}_{i} \gets \log(1+\boldsymbol 1/\boldsymbol{\alpha}_i)$ \Comment{entrywise division}
    \State $\tilde{\boldsymbol y}_{i} \gets \log\boldsymbol\alpha_{i}-\tilde{\boldsymbol\sigma}^{2}_{i}/2$
    \Comment{log-space pseudo-observation}
\EndFor
\For{$j=1,\ldots,K$}
    \State $\tilde{\boldsymbol y}^{(j)}\gets (\tilde y_{1j},\ldots,\tilde y_{Nj})^\top$
    \State $\boldsymbol\Lambda_j\gets \operatorname{diag}(\tilde\sigma_{1j}^2,\ldots,\tilde\sigma_{Nj}^2)$
    \State $f_j(\cdot)\sim\mathcal{GP}(0,K_\theta)$
\EndFor
\State Optimize shared $\theta$ by maximizing
\Statex $\sum_{j=1}^{K}\log\mathcal N\!\left(\tilde{\boldsymbol y}^{(j)}\mid \mathbf 0,\mathbf K_{\theta}(\mathbf X,\mathbf X)+\boldsymbol\Lambda_j\right)$
\State Predict by sampling $\mathbf f_*$ from the GP posterior and averaging $\mathrm{softmax}(\mathbf f_*)$
\end{algorithmic}
\vspace{0.35em}
\hrule
\end{minipage}
Some sparse experiments add a learned per-output noise scale to the fixed likelihood variance. This changes only the diagonal covariance used in the Gaussian marginal likelihood or its sparse lower bound; it does not change the pseudo-observations $\boldsymbol z_i$ or $\tilde{\boldsymbol y}_i$ shown above.

\section{Additional Experimental Details and Results}\label{app:exp}

\paragraph{UCI experiments}
Table~\ref{tab:uci} shows the number of training and test points, the number of categories and the number of inducing points for the experiments considered from the UCI repository \citep{Dua:2019}, the last three rows use the full training data in the model fitting and do not need inducing points. 
Table~\ref{tab:uci_metrics} reports, for each UCI dataset and model, the hyperparameters selected via validation and the values of ACC, NLL and ECE on test. Each experiment was run for 5 random seeds, for each seed we fit the kernel parameters with marginal likelihood or its lower bound, and we selected the best hyperparmeters (normalization, learning rate, $\alpha_\varepsilon$ and $\lambda$) based on the validation loss of the individual seed. The validation loss taken is $\tfrac{1}{2}(\mathrm{NLL}_{\mathrm{val}} + \mathrm{ECE}_{\mathrm{val}})$. The best result per experiment (mean over five random seeds) is highlighted in bold.
These are the same values shown in the exact-setting and sparse-setting summaries in Figures~\ref{fig:uci_conjugate} and~\ref{fig:uci_sparse}.

The code for our method was implemented using \texttt{GPyTorch} \citet{gardner2018gpytorch} and will be made publicly available upon acceptance. The auxiliary-variable methods (LSM and BSM) were run using the Julia packages \texttt{AugmentedGaussianProcesses.jl} and \texttt{AugmentedGPLikelihoods.jl} respectively.

\begin{table}[ht]
\centering
\caption{Details on all datasets from the UCI repository used in the experiments.}
\begin{tabular}{lccccccc}
\hline
Dataset & \# Instances & Training Instances & Test Instances & \# Attributes & \# Classes & Inducing Points & Source \\
\hline
EEG      & 14980  & 10980  & 4000  & 14 & 2  & 200 & UCI \\
HTRU2    & 17898  & 12898  & 5000  & 8  & 2  & 200 & UCI \\
MAGIC    & 19020  & 14020  & 5000  & 10 & 2  & 200 & UCI \\
MINIBOO  & 130064 & 120064 & 10000 & 50 & 2  & 400 & UCI \\
LETTER   & 20000  & 15000  & 5000  & 16 & 26 & 200 & UCI \\
DRIVE    & 58509  & 48509  & 10000 & 48 & 11 & 500 & UCI \\
MOCAP    & 78095  & 68095  & 10000 & 37 & 5  & 500 & UCI \\
\hline
New-thyroid & 215 & 165 & 50 & 5  & 3 & -- & UCI \\
Wine        & 178 & 128 & 50 & 13 & 3 & -- & UCI \\
Glass       & 214 & 164 & 50 & 9  & 6 & -- & UCI \\
\hline
\end{tabular}
\label{tab:uci}
\end{table}

\begin{figure}
    \centering
    \includegraphics[width=0.2\linewidth]{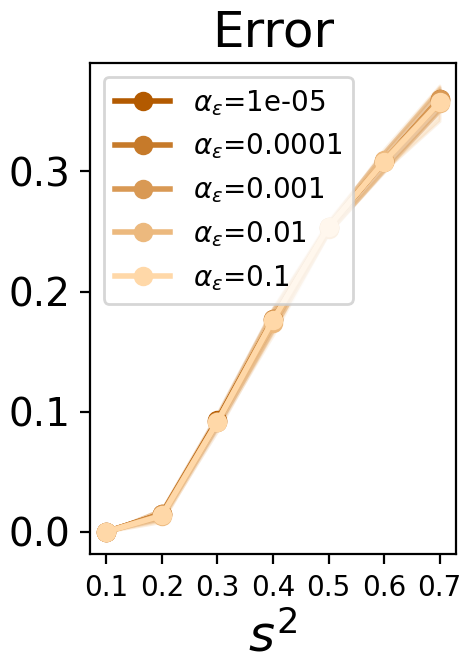}
    \includegraphics[width=0.2\linewidth]{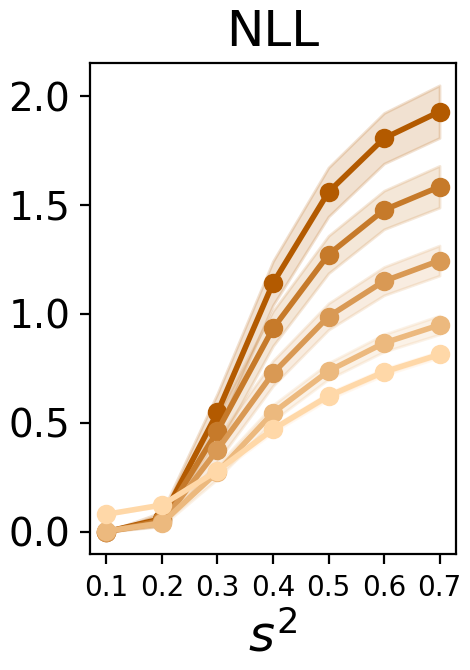}
    \includegraphics[width=0.2\linewidth]{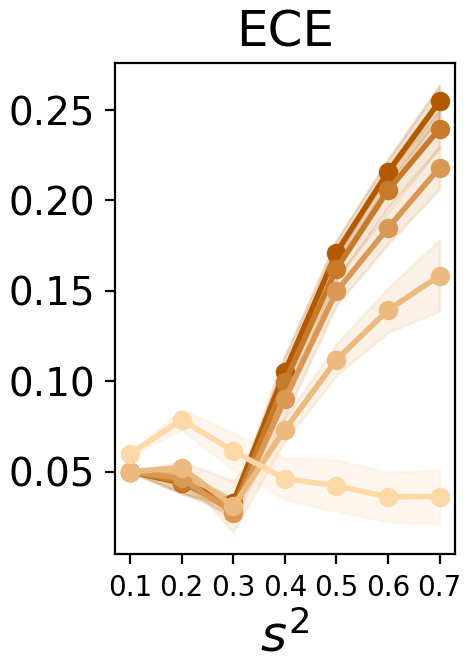}
    \caption{Error, NLL and ECE for increasing overlap of the input variables for GPD and different values of $\alpha_\varepsilon$. }
    \label{fig:toy_gpd}
\end{figure}

\paragraph{The effect of the parameter $\alpha_{\varepsilon}$}
Following the same setup as in the experiment ``Effect of the parameter $\lambda$'' we study the effect of $\alpha_{\varepsilon}$ in GPD as the class assignments become less clear for $s\in\{0.1,..,0.7\}$. Figure~\ref{fig:toy_gpd} shows a similar behavior as we found for $\lambda$. Values closer to zero work well when the classes are well separated, but values further from zero work better when the classes are intertwined and remain accurate and well calibrated when one input value can have multiple classes. The choice of $\alpha_{\varepsilon}$ is problem dependent and $K$-dependent, and the authors of GPD suggest it be chosen by validation or cross-validation. 

\begin{figure}
    \centering
    \includegraphics[width=0.2\linewidth]{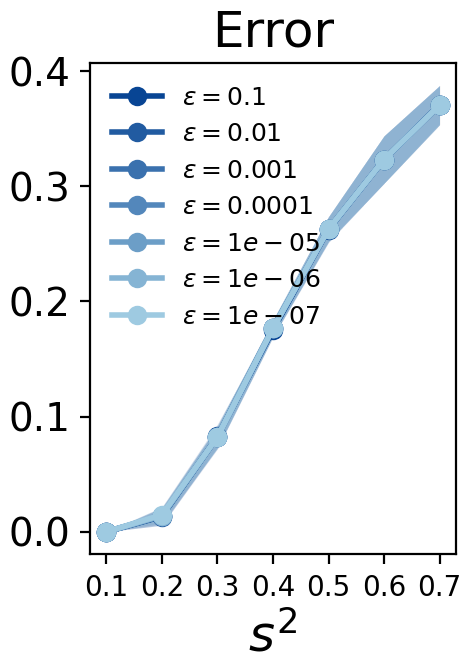}
    \includegraphics[width=0.2\linewidth]{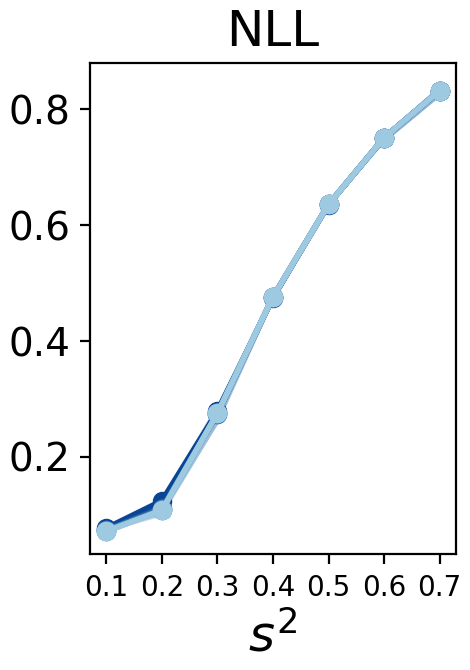}
    \includegraphics[width=0.2\linewidth]{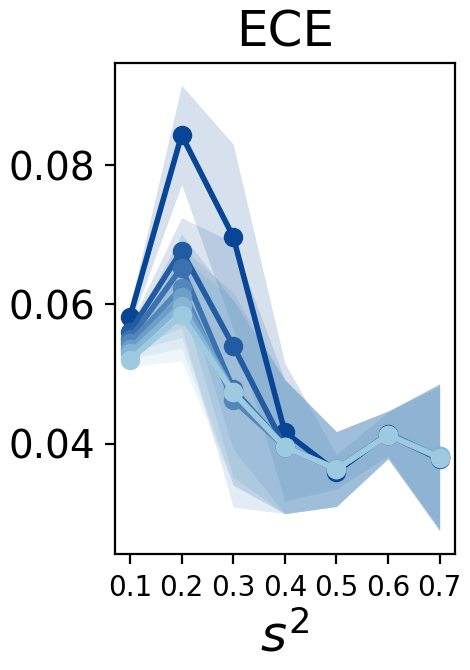}
    \caption{
    Error, NLL and ECE for increasing overlap of the input variables for different values of the error tolerance $\varepsilon$ and $\lambda=0.9$.
    }
    \label{fig:sigma}
\end{figure}

\paragraph{The effect of the parameter $\sigma$}
For the same setup as ``Effect of the parameter $\lambda$'', we fix $\lambda=0.9$ in \ourmethod\ and and vary $\varepsilon\in \{0.1,.., 10^{-7}\}$. The parameter $\sigma$ is set based the lower bound dependent on $\varepsilon$ as in Proposition~\ref{prop:sigma_bound}. We can see in  Fig.~\ref{fig:sigma} that for smaller tolerance we obtain better calibration.
This motivates choosing a small value for the tolerance such as $10^{-6}$. 

\begin{figure}
    \centering
    \includegraphics[width=0.5\linewidth]{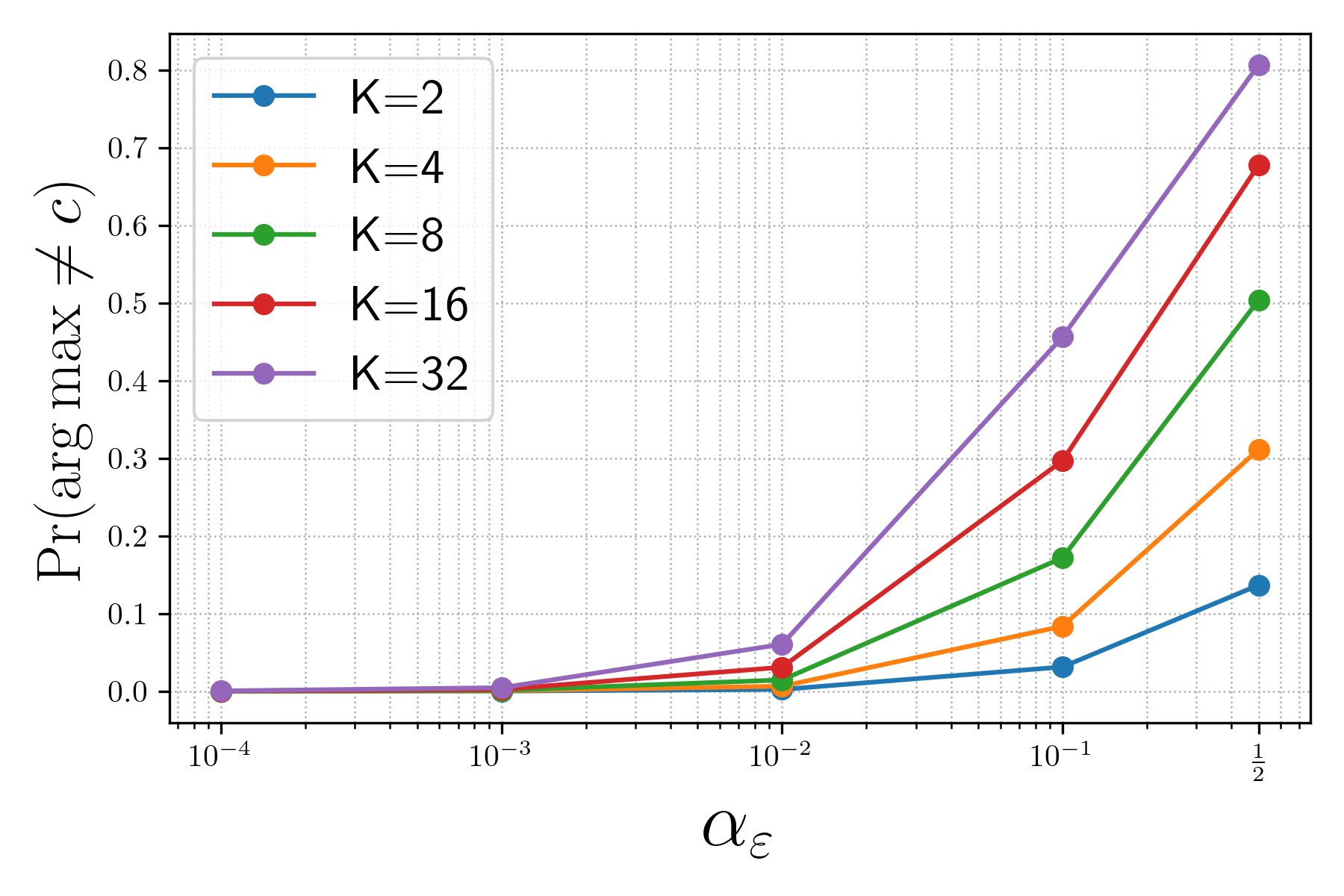}
    \caption{The error rate of GDP for the estimate $\hat{\boldsymbol{\pi}}= \mathrm{softmax}(\boldsymbol{z} )$ with $\boldsymbol z\sim \mathrm{Lognormal}(\tilde{y},\tilde{\sigma}^2)$. This error is due to the heavy tails of the Lognormal distribution that fall on the $\arg\max$ regions of the other categories. }
    \label{fig:gpd_error}
\end{figure}

\paragraph{Error rate of GPD}
Here we analyze the behavior of the error rate of GPD when using the estimates  $\hat{\boldsymbol{\pi}}= \mathrm{softmax}(\boldsymbol{z} )$ with $\boldsymbol z\sim \mathrm{Lognormal}(\tilde{y},\tilde{\sigma}^2)$. There is no need to introduce a GP, we just analyze how often the samples from the Lognormal distribution fall outside the intended $\arg\max$ region with the parameter construction presented in Eq.~\ref{eq:lognorm}. 
Figure~\ref{fig:gpd_error} shows the error rate of recovering the true category as a function of the number of classes and the value of $\alpha_{\varepsilon}$. 
The error estimates are computed  with $100,000$ samples of $\boldsymbol{\pi}$ each using a single draw of $\boldsymbol{z}$. Empirically we see that  error decreases for values of $\alpha_{\varepsilon}$ closer to zero, and the error increases for larger number of categories. 

\begin{figure}
    \centering
    \includegraphics[width=0.5\linewidth]{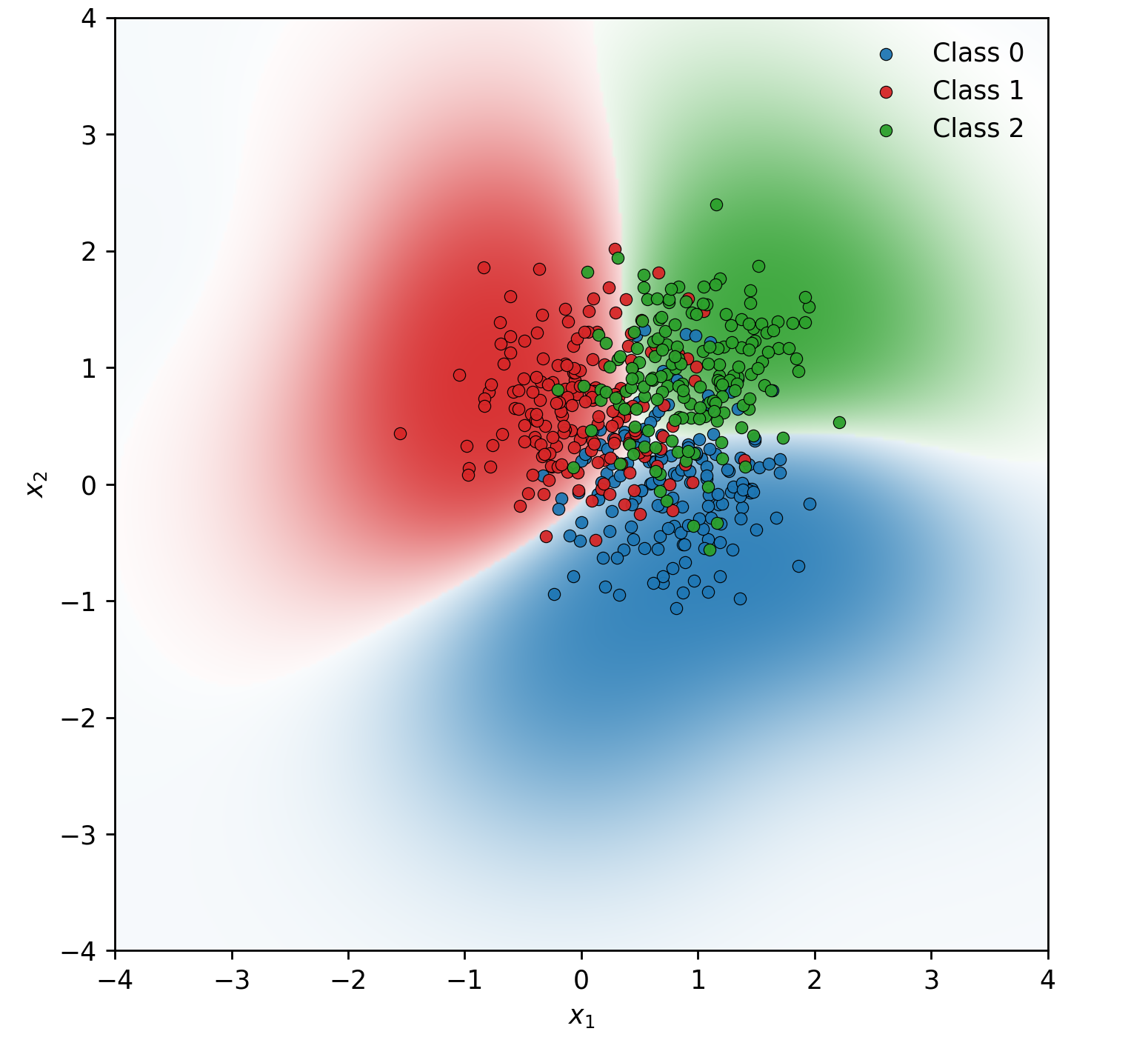}
    \caption{    
    Training points are shown with their true class labels. The background color indicates the predicted class, and its intensity of the color indicates the predicted probability. 
    The method handles out of distribution data by reverting to uniform probabilities over classes. This can be seen by the probabilities smoothly transitioning to the uniform distribution (white) far from the data.
 }
    \label{fig:decision_boundaries}
\end{figure}

\paragraph{Visualization of decision boundaries}
Visualization of the Exact-ILR predictive probabilities on the same setup used in the experiment “Effect of the parameter $\lambda$”, with $\lambda=0.9$ and $s^2=0.5$. 
Figure~\ref{fig:decision_boundaries} shows smooth decision boundaries between the classes, with predicted probabilities that transition smoothly toward the uniform distribution both far from the data and near the decision boundaries. In this simple experiment, the method exhibits well-behaved and smooth estimates.

\begin{table}[ht]
\centering
\caption{Full details on the hyperparameters selected for each model and experiment. Accuracy, NLL and ECE are computed on the test dataset, and the mean best value is boldfaced. Norm. refers to the input normalization method, HP to hyperparameter, and LR to the learning rate.  }
\scriptsize 
\begin{tabular}{lllllllll}
\toprule
Exp & Model & Norm. & Target HP & LR & ACC $\uparrow$ & NLL $\downarrow$ & ECE $\downarrow$ & Fit time (s) \\
\midrule
Wine & LSM & $[-1,1]$ & -- & 0.001 & 0.98 $\pm$ 0.017 & 0.52 $\pm$ 0.023 & 0.36 $\pm$ 0.010 & -- \\
 & BSM & $[-1,1]$ & -- & 0.001 & 0.97 $\pm$ 0.023 & 0.87 $\pm$ 0.011 & 0.55 $\pm$ 0.024 & -- \\
 & GPD & $[-1,1]$, Z & $\alpha_\varepsilon$: 0.0001 & 0.01 & \textbf{0.98 $\pm$ 0.017} & 0.06 $\pm$ 0.021 & 0.07 $\pm$ 0.017 & 5.9 \\
 & E-ILR (ours) & $[-1,1]$, Z & $\lambda$: 0.999999 & 0.01 & \textbf{0.98 $\pm$ 0.017} & \textbf{0.06 $\pm$ 0.018} & \textbf{0.07 $\pm$ 0.018} & 6.2 \\
\midrule
Glass & LSM & $[-1,1]$ & -- & 0.001 & 0.56 $\pm$ 0.041 & 1.24 $\pm$ 0.028 & 0.26 $\pm$ 0.049 & -- \\
 & BSM & $[-1,1]$ & -- & 0.001 & 0.66 $\pm$ 0.088 & 1.53 $\pm$ 0.010 & 0.42 $\pm$ 0.087 & -- \\
 & GPD & Z & $\alpha_\varepsilon$: 0.01, 0.0001, 0.001 & 0.01, 0.001 & 0.70 $\pm$ 0.059 & 0.90 $\pm$ 0.157 & \textbf{0.13 $\pm$ 0.041} & 7.4 \\
 & E-ILR (ours) & Z, $[-1,1]$ & \makecell[l]{$\lambda$: 0.99, 0.999999, \\0.999, 0.9999, 0.99999} & 0.001, 0.01 & \textbf{0.71 $\pm$ 0.061} & \textbf{0.82 $\pm$ 0.137} & 0.13 $\pm$ 0.028 & 7.7 \\
\midrule
Thyroid & LSM & $[-1,1]$ & -- & 0.001 & 0.83 $\pm$ 0.046 & 0.44 $\pm$ 0.042 & 0.14 $\pm$ 0.041 & -- \\
 & BSM & $[-1,1]$ & -- & 0.001 & 0.82 $\pm$ 0.052 & 0.83 $\pm$ 0.017 & 0.39 $\pm$ 0.047 & -- \\
 & GPD & $[-1,1]$ & $\alpha_\varepsilon$: 0.01, 0.001 & 0.001, 0.01 & 0.94 $\pm$ 0.033 & \textbf{0.17 $\pm$ 0.111} & 0.06 $\pm$ 0.012 & 6.4 \\
 & E-ILR (ours) & $[-1,1]$, Z & \makecell[l]{$\lambda$: 0.999, 0.9999,\\ 0.99999, 0.999999} & 0.001, 0.01 & \textbf{0.96 $\pm$ 0.026} & 0.18 $\pm$ 0.106 & \textbf{0.05 $\pm$ 0.020} & 5.9 \\
\midrule
EEG & LSM & Z & -- & 0.01 & 0.65 $\pm$ 0.047 & 0.60 $\pm$ 0.027 & 0.07 $\pm$ 0.016 & -- \\
 & BSM & Z, $[-1,1]$ & -- & 0.01 & 0.79 $\pm$ 0.018 & 0.51 $\pm$ 0.024 & 0.13 $\pm$ 0.010 & -- \\
 & GPD & Z & $\alpha_\varepsilon$: 0.01 & 0.01 & 0.88 $\pm$ 0.034 & 0.28 $\pm$ 0.077 & 0.03 $\pm$ 0.004 & 108 \\
 & SVGPC & Z & -- & 0.01 & 0.81 $\pm$ 0.027 & 0.40 $\pm$ 0.044 & 0.04 $\pm$ 0.007 & 29.4 \\
 & U-ILR (ours) & Z & -- & 0.01 & 0.82 $\pm$ 0.026 & 0.39 $\pm$ 0.046 & 0.03 $\pm$ 0.005 & 29.8 \\
 & C-ILR (ours) & Z & $\lambda$: 0.99 & 0.001 & \textbf{0.89 $\pm$ 0.034} & \textbf{0.26 $\pm$ 0.070} & \textbf{0.03 $\pm$ 0.008} & 56.5 \\
\midrule
HTRU2 & LSM & $[-1,1]$ & -- & 0.01 & 0.97 $\pm$ 0.002 & 0.09 $\pm$ 0.004 & 0.05 $\pm$ 0.002 & -- \\
 & BSM & Z, $[-1,1]$ & -- & 0.01, 0.001 & 0.98 $\pm$ 0.002 & 0.40 $\pm$ 0.002 & 0.31 $\pm$ 0.002 & -- \\
 & GPD & $[-1,1]$ & $\alpha_\varepsilon$: 0.01 & 0.01 & 0.98 $\pm$ 0.002 & 0.08 $\pm$ 0.006 & \textbf{0.04 $\pm$ 0.001} & 92.4 \\
 & SVGPC & $[-1,1]$, Z & -- & 0.01 & 0.98 $\pm$ 0.002 & 0.07 $\pm$ 0.005 & 0.04 $\pm$ 0.001 & 25.9 \\
 & U-ILR (ours) & Z, $[-1,1]$ & -- & 0.01 & 0.98 $\pm$ 0.002 & \textbf{0.07 $\pm$ 0.004} & 0.04 $\pm$ 0.001 & 32.2 \\
 & C-ILR (ours) & Z, $[-1,1]$ & $\lambda$: 0.99 & 0.01 & \textbf{0.98 $\pm$ 0.002} & 0.07 $\pm$ 0.005 & 0.04 $\pm$ 0.002 & 52.3 \\
\midrule
Magic & LSM & $[-1,1]$ & -- & 0.01 & 0.82 $\pm$ 0.002 & 0.43 $\pm$ 0.003 & 0.10 $\pm$ 0.005 & -- \\
 & BSM & $[-1,1]$, Z & -- & 0.01 & 0.86 $\pm$ 0.002 & 0.45 $\pm$ 0.002 & 0.17 $\pm$ 0.002 & -- \\
 & GPD & Z & $\alpha_\varepsilon$: 0.1 & 0.01 & 0.87 $\pm$ 0.003 & 0.33 $\pm$ 0.004 & 0.03 $\pm$ 0.002 & 126 \\
 & SVGPC & $[-1,1]$, Z & -- & 0.01 & 0.86 $\pm$ 0.003 & 0.33 $\pm$ 0.004 & 0.03 $\pm$ 0.003 & 29.1 \\
 & U-ILR (ours) & Z & -- & 0.01 & 0.86 $\pm$ 0.003 & 0.33 $\pm$ 0.003 & 0.02 $\pm$ 0.004 & 25.3 \\
 & C-ILR (ours) & Z & $\lambda$: 0.95 & 0.01 & \textbf{0.87 $\pm$ 0.002} & \textbf{0.32 $\pm$ 0.005} & \textbf{0.02 $\pm$ 0.006} & 55.5 \\
\midrule
MiniBoo & LSM & Z & -- & 0.01 & 0.89 $\pm$ 0.003 & 0.27 $\pm$ 0.003 & 0.07 $\pm$ 0.004 & -- \\
 & BSM & Z & -- & 0.01 & 0.91 $\pm$ 0.002 & 0.26 $\pm$ 0.003 & 0.08 $\pm$ 0.003 & -- \\
 & GPD & Z & $\alpha_\varepsilon$: 0.1 & 0.001 & 0.91 $\pm$ 0.003 & 0.23 $\pm$ 0.005 & 0.04 $\pm$ 0.003 & 5587 \\
 & SVGPC & Z & -- & 0.01 & 0.90 $\pm$ 0.001 & 0.23 $\pm$ 0.003 & 0.03 $\pm$ 0.001 & 45.8 \\
 & U-ILR (ours) & Z & -- & 0.01 & 0.90 $\pm$ 0.001 & 0.23 $\pm$ 0.003 & \textbf{0.03 $\pm$ 0.002} & 42.8 \\
 & C-ILR (ours) & Z & $\lambda$: 0.95 & 0.01, 0.001 & \textbf{0.91 $\pm$ 0.005} & \textbf{0.22 $\pm$ 0.010} & 0.03 $\pm$ 0.002 & 2535 \\
\midrule
Drive & LSM & $[-1,1]$ & -- & 0.01, 0.001 & 0.76 $\pm$ 0.014 & 0.69 $\pm$ 0.066 & 0.23 $\pm$ 0.022 & -- \\
 & BSM & $[-1,1]$ & -- & 0.01 & 0.93 $\pm$ 0.010 & 0.54 $\pm$ 0.006 & 0.32 $\pm$ 0.007 & -- \\
 & GPD & $[-1,1]$ & $\alpha_\varepsilon$: 0.0001 & 0.01 & \textbf{0.99 $\pm$ 0.001} & \textbf{0.02 $\pm$ 0.002} & \textbf{0.05 $\pm$ 0.000} & 16839 \\
 & SVGPC & $[-1,1]$ & -- & 0.01 & 0.99 $\pm$ 0.002 & 0.03 $\pm$ 0.004 & 0.05 $\pm$ 0.001 & 492 \\
 & U-ILR (ours) & $[-1,1]$ & -- & 0.01 & 0.99 $\pm$ 0.001 & 0.03 $\pm$ 0.003 & 0.05 $\pm$ 0.001 & 495 \\
 & C-ILR (ours) & $[-1,1]$ & $\lambda$: 0.999999 & 0.01 & 0.99 $\pm$ 0.001 & 0.02 $\pm$ 0.002 & 0.05 $\pm$ 0.000 & 16099 \\
\midrule
letter & LSM & $[-1,1]$ & -- & 0.01 & 0.76 $\pm$ 0.004 & 1.47 $\pm$ 0.009 & 0.48 $\pm$ 0.004 & -- \\
 & BSM & Z & -- & 0.01 & 0.78 $\pm$ 0.008 & 0.91 $\pm$ 0.005 & 0.28 $\pm$ 0.008 & -- \\
 & GPD & $[-1,1]$ & $\alpha_\varepsilon$: 0.0001, 0.001 & 0.01 & \textbf{0.96 $\pm$ 0.002} & \textbf{0.12 $\pm$ 0.003} & 0.04 $\pm$ 0.004 & 4443 \\
 & SVGPC & $[-1,1]$ & -- & 0.01 & 0.96 $\pm$ 0.002 & 0.13 $\pm$ 0.008 & \textbf{0.04 $\pm$ 0.002} & 697 \\
 & U-ILR (ours) & $[-1,1]$ & -- & 0.01 & 0.96 $\pm$ 0.002 & 0.13 $\pm$ 0.008 & 0.04 $\pm$ 0.003 & 661 \\
 & C-ILR (ours) & $[-1,1]$ & $\lambda$: 0.999999 & 0.01 & 0.95 $\pm$ 0.002 & 0.16 $\pm$ 0.003 & 0.04 $\pm$ 0.001 & 4491 \\
\midrule
MoCap & LSM & Z, $[-1,1]$ & -- & 0.01 & 0.88 $\pm$ 0.062 & 0.58 $\pm$ 0.482 & 0.22 $\pm$ 0.146 & -- \\
 & BSM & $[-1,1]$ & -- & 0.01 & 0.94 $\pm$ 0.001 & 0.50 $\pm$ 0.002 & 0.28 $\pm$ 0.001 & -- \\
 & GPD & $[-1,1]$ & $\alpha_\varepsilon$: 0.001 & 0.01 & 0.99 $\pm$ 0.001 & 0.05 $\pm$ 0.005 & 0.05 $\pm$ 0.001 & 5497 \\
 & SVGPC & $[-1,1]$ & -- & 0.01 & 0.97 $\pm$ 0.001 & 0.11 $\pm$ 0.005 & \textbf{0.04 $\pm$ 0.002} & 154 \\
 & U-ILR (ours) & $[-1,1]$ & -- & 0.01 & 0.97 $\pm$ 0.001 & 0.11 $\pm$ 0.003 & 0.05 $\pm$ 0.002 & 109 \\
 & C-ILR (ours) & $[-1,1]$ & $\lambda$: 0.999 & 0.01 & \textbf{0.99 $\pm$ 0.001} & \textbf{0.05 $\pm$ 0.005} & 0.05 $\pm$ 0.001 & 2738 \\
\bottomrule
\end{tabular}

\label{tab:uci_metrics}
\end{table}

\end{document}

%% file: commands.tex
\usepackage{amsmath,amsfonts,bm,amsthm}
\usepackage{amsthm}

\def\1{\bm{1}}

\DeclareMathAlphabet{\mathsfit}{\encodingdefault}{\sfdefault}{m}{sl}
\SetMathAlphabet{\mathsfit}{bold}{\encodingdefault}{\sfdefault}{bx}{n}

\newcommand{\ourmethod}{Exact-ILR}
\newcommand{\ourmethodsparse}{Collapsed-ILR}
\newcommand{\ourmethodvariational}{Uncollapsed-ILR}

\newcommand{\compmethodb}{SVGPC}